\documentclass[twocolumn,10pt]{asme2ej}

\usepackage{graphicx} 
\newcommand{\EQ}{\begin{equation}}
\newcommand{\NQ}{\end{equation}}
\newcommand{\ER}{\begin{eqnarray}}
\newcommand{\NR}{\end{eqnarray}}
\newcommand{\ERS}{\begin{eqnarray*}}
\newcommand{\NRS}{\end{eqnarray*}}
\newcommand{\bit}{\begin{itemize}}
\newcommand{\ben}{\begin{enumerate}}
\newcommand{\eben}{\end{enumerate}}
\newcommand{\ebit}{\end{itemize}}
\newcommand{\bzero}{{\bf 0}}
\newcommand{\ba}{{\bf a}}

\newcommand{\bc}{{\bf c}}

\newcommand{\bh}{{\bf h}}

\newcommand{\bm}{{\bf m}}

\newcommand{\bq}{{\bf q}}

\newcommand{\bu}{{\bf u}}
\newcommand{\bv}{{\bf v}}

\newcommand{\bx}{{\bf x}}
\newcommand{\by}{{\bf y}}

\newcommand{\bA}{{\bf A}}
\newcommand{\bB}{{\bf B}}

\newcommand{\bF}{{\bf F}}

\newcommand{\bI}{{\bf I}}
\newcommand{\bJ}{{\bf J}}
\newcommand{\bK}{{\bf K}}

\newcommand{\bM}{{\bf M}}

\newcommand{\bP}{{\bf P}}
\newcommand{\bQ}{{\bf Q}}

\newcommand{\dq}{\dot q}

\newcommand{\dtheta}{\dot \theta}

\def\m@th{\mathsurround=0pt }

\def\eqalign#1{\null\,\vcenter{\openup1\jot \m@th
  \ialign{\strut\hfil$\displaystyle{##}$&$\displaystyle{{}##}$\hfil
      \crcr#1\crcr}}\,}

\def\displ@y{\global\dt@ptrue \openup1\jot \m@th
  \everycr{\noalign{\ifdt@p \global\dt@pfalse
      \vskip-\lineskiplimit \vskip\normallineskiplimit
      \else \penalty\interdisplaylinepenalty \fi}}}
\def\@lign{\tabskip=0pt\everycr={}} 

\def\eqalignno#1{\displ@y \tabskip=\centering
  \halign to\displaywidth{\hfil$\@lign\displaystyle{##}$\tabskip=0pt
    &$\@lign\displaystyle{{}##}$\hfil\tabskip=\centering
    &\llap{$\@lign##$}\tabskip=0pt\crcr
    #1\crcr}}

\def\oalpha{{\ooalign{\hfil\raise.07ex\hbox{$\alpha$}\hfil\crcr\mathhexbox20D}}}

\def\ooalpha{{\ooalign{\hfil\raise.07ex\hbox{$\alpha$}\hfil\crcr$\textstyle{\circ}t$}}}

\def\obeta{{\ooalign{\hfil\raise.07ex\hbox{$\scriptstyle{\beta}$}\hfil\crcr$\textstyle{\circ}$}}}

\def\ot{{\ooalign{\hfil\raise.07ex\hbox{$t$}\hfil\crcr$\textstyle{\circ}$}}}

\def\oooalpha{{\ooalign{\hfil\raise.07ex\hbox{${\scriptstyle{\alpha}}$}\hfil\crcr${\textstyle{\circ}}$}}}

\def\trademark{{\ooalign{\hfil\raise.07ex\hbox{R}\hfil\crcr\mathhexbox20D}}}

\usepackage{graphicx}          

\usepackage{mathptmx} 
\usepackage{graphicx,wrapfig,amsmath,hhline,epsfig,verbatim,url,amssymb,cite}

\usepackage{times,color,soul} 

\usepackage{mathptmx} 
\usepackage{bm}
\usepackage{float}
\usepackage{color,soul}
\usepackage{dsfont}
\usepackage{comment}
\usepackage{algorithm}
\usepackage{algorithmic}
\usepackage{psfrag}   
\usepackage{url}
\usepackage{hyperref}
\usepackage{setspace}
\usepackage[table]{xcolor}
\usepackage{paralist}
\usepackage{booktabs}

\usepackage[switch]{lineno}
\linenumbers

\usepackage{flushend}

\usepackage{pifont}

\newtheorem{defn}{Definition}
\newtheorem{thm}{Theorem}
\newtheorem{rem}{Remark}
\newtheorem{prop}{Proposition}

\title{Global-Position Tracking Control of 3-D Bipedal Walking via Virtual Constraint Design and Multiple Lyapunov Analysis}

\author{Yan Gu 
\thanks{Corresponding Author. This material is based upon work supported by the National Science Foundation under Grant No. CMMI-1934280.
	Any opinions, findings, and conclusions or recommendations expressed in this material are those of the author(s) and do not necessarily reflect the views of the National Science Foundation.} 
    \affiliation{
	Department of Mechanical Engineering\\
	University of Masachusetts Lowell\\
	Lowell, MA 01854, U.S.A.\\
    Email: yan\_gu@uml.edu
    }	
}

\author{Yuan Gao 
    \affiliation{
    Department of Mechanical Engineering\\
	University of Masachusetts Lowell\\
	Lowell, MA 01854, U.S.A.\\
    Email: yuan\_gao@student.uml.edu
    }
}

\author{Bin Yao\\
        School of Mechanical Engineering\\
        Purdue University\\
        West Lafayette, IN 47907, U.S.A.\\
        Email: byao@purdue.edu
        \\
        \\
       {\tensfb C. S. George Lee}
    \affiliation{
        School of Electrical and Computer Engineering\\
        Purdue University\\
        West Lafayette, IN 47907, U.S.A.\\
        Email: csglee@purdue.edu
    }
}

\begin{document}

\maketitle    

\begin{abstract}
A safety-critical measure of legged locomotion performance is a robot's ability to track its desired time-varying position trajectory in an environment, which is herein termed as ``global-position tracking".
This paper introduces a nonlinear control approach that achieves asymptotic global-position tracking for three-dimensional (3-D) bipedal robot walking.
Designing a global-position tracking controller presents a challenging problem due to the complex hybrid robot model
and the time-varying desired global-position trajectory.
Towards tackling this problem, 
the first main contribution is the construction of impact invariance to ensure all desired trajectories respect the foot-landing impact dynamics, which is a necessary condition for realizing asymptotic tracking of hybrid walking systems.
Thanks to their independence of the desired global position,
these conditions can be exploited to decouple the higher-level planning of the global position and the lower-level planning of the remaining trajectories,
thereby greatly alleviating the computational burden of motion planning.
The second main contribution is the Lyapunov-based stability analysis of the hybrid closed-loop system, which produces sufficient conditions to guide the controller design for achieving asymptotic global-position tracking during fully actuated walking.
Simulations and experiments on a 3-D bipedal robot with twenty revolute joints confirm the validity of the proposed control approach in guaranteeing accurate tracking.
\end{abstract}

\vspace{0in}
\begin{figure*}[t]
	\centering
	\includegraphics[width=1\linewidth]{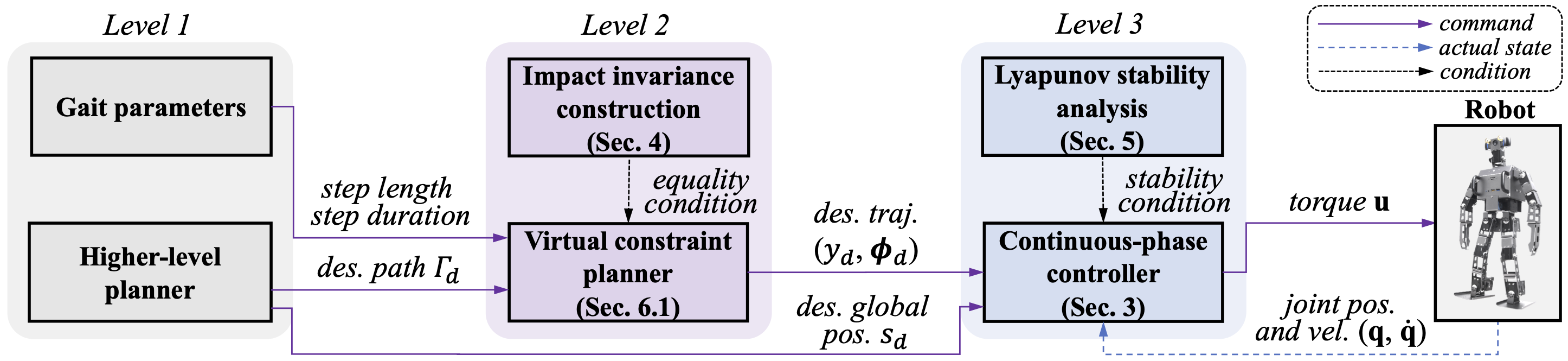}
	\caption{
		{{Overview of the proposed control approach. This study focuses on impact invariance construction in Level 2 and stability analysis and controller design in Level 3.}}
	}
	\label{overview}
\end{figure*}
\vspace{0in}

\section{Introduction}
\label{sec-intro}

A robot's global position represents its absolute position in an environment. 
Poor global-position tracking can potentially put the safety of both humans and robots at risk, for example, by causing robots' failure to avoid pedestrians in human-populated environments. 
To achieve accurate global-position tracking, the Zero-Moment-Point (ZMP) control approach has been introduced based on the ZMP balance criterion and the continuous-time dynamic model of bipedal walking~\cite{vukobratovic2001zero,kajita2003biped,kim2006experimental}.
{Yet, bipedal walking is inherently a hybrid process involving both continuous motions (e.g., foot swinging) and discrete impact dynamics (e.g., sudden joint-velocity jumps upon a foot landing)~\cite{golliday1977approach, hurmuzlu1986role,bhounsule2018switching,yeatman2019decentralized}.}
Achieving reliable global-position tracking by explicitly addressing the hybrid robot dynamics presents substantial challenges.

This study focuses on addressing the challenges associated with: a) lower-level trajectory generation (i.e., Level 2 in Fig.~\ref{overview}) and b) controller design (i.e., Level 3 in Fig.~\ref{overview}).
It is assumed that
the desired global path and the desired time-varying position trajectory along the path have both been provided by a higher-level planner (i.e., Level 1 in Fig.~\ref{overview}) without impact dynamics considered.

A key challenge in the lower-level trajectory generation (i.e., Level 2 in {Fig.~\ref{overview}}) is to reduce the computational burden caused by enforcing the impact dynamics on the desired trajectories.
For the controller to achieve asymptotic tracking based on a hybrid robot model, the desired trajectories need to agree with the impact dynamics; i.e., their pre- and post-impact values should satisfy the impact map.
This is because the impact dynamics cannot be directly controlled due to their infinitesimally short duration
~\cite{gu2017time,rijnen2019hybrid,rijnen2019sensitivity,wang2020impact}.
Yet, the computational burden caused by respecting the impact dynamics is heavy because of the high dimension of a legged robot's state space and the nonlinearity of the impact dynamics.

Another challenge is the closed-loop stability analysis of the hybrid dynamical system that produces sufficient conditions to inform the controller derivation (i.e., Level 3 in {Fig.~\ref{overview}}).
Such a stability analysis is complex because a closed-loop system capable of stabilizing a time-varying global-position trajectory is hybrid, nonlinear, and time-varying with uncontrolled, state-triggered impact dynamics.

\label{sec-related}

\subsection{Related Work on Orbitally Stabilizing Control}

The most widely studied control approach that explicitly addresses the hybrid walking dynamics is the Hybrid Zero Dynamics (HZD) method~\cite{grizzle2001asymptotically,westervelt2003hybrid,morris2009hybrid,martin2015hybrid,gong2019feedback,fevre2019terrain}.
The HZD method provably stabilizes dynamic walking motions through orbital stabilization of the hybrid closed-loop control system.
It has realized remarkable performance for various gait types such as periodic underactuated~\cite{hamed2014event,da2016from},
fully actuated~\cite{ames2012dynamically}, and multi-domain walking~\cite{akbari2019dynamic}.

The HZD framework introduces virtual constraints to represent the evolution of a robot's desired configuration with respect to a phase variable that indicates how far a step has progressed.
To enforce the impact dynamics on the desired gait,
the HZD approach introduces a method termed ``impact invariance construction'' to produce an equality constraint under which the desired gait respects the impact dynamics, and incorporates the constraint in the optimization-based generation of virtual constraints.
Yet, because the encoding of the global-position trajectory is inherently different from that of virtual constraints, the previous impact invariance construction cannot be directly applied or extended to ensure the agreement with impact dynamics for the desired global-position trajectory.
Specifically, the virtual constraints are encoded by a local phase variable that is reset at the beginning of a walking step,
while the desired global-position trajectory is usually encoded by a global phase variable that involves continuously and monotonically across all walking steps.

To analyze the closed-loop stability for guiding controller designs, the HZD approach exploits the Poincar\'e section method to examine the asymptotic convergence of a robot's state to the desired periodic orbit representing the desired gait in the state space.
Recently, the HZD framework has been extended to achieve asymptotic tracking of the desired global path during 3-D underactuated bipedal walking~\cite{xiong2020global}.
Yet, an orbitally stabilizing controller cannot stabilize a prespecified time-varying trajectory~\cite{khalil1996nonlinear} such as the desired global-position trajectory.

\subsection{Related Work on Trajectory Tracking Control}

Our previous trajectory tracking controller designs either focus on individual joint trajectory tracking~\cite{gu2020adaptive,gu2021adaptive} or only considers 2-D walking~\cite{gu2016bipedal,gu2017exponential,gao2019global}.
In particular, our previous work on 2-D walking, including the impact invariance construction and stability analysis, is not valid for 3-D walking.
Specifically, robot dynamics during 3-D walking are nonlinearly coupled in the heading and lateral directions of the robot's global path, but 2-D walking does not exhibit lateral motion, and accordingly, the coupling is trivial.
This nonlinear coupling significantly increases the complexity of controller derivation in addressing 3-D walking compared with 2-D walking.
Furthermore, experimental validation of these previous controllers has been missing.

Beyond the scope of global-position tracking control for bipedal walking robots,
\textcolor{black}{trajectory tracking control}
of general hybrid systems with state-triggered jumps is an active research topic ~\cite{menini2001asymptotic,biemond2012tracking,forni2013follow,naldi2013passivity,rijnen2016hybrid,rijnen2017control}.
Lyapunov-based controller design methodologies have been introduced to provably achieve asymptotic trajectory tracking {for linear hybrid systems}~\cite{biemond2012tracking,forni2013follow}.
In this study, to guide the needed controller design, we will extend the previous Lyapunov-based stability analysis to nonlinear hybrid systems that include 3-D bipedal robot walking.

\subsection{Contributions}

This study aims to derive and experimentally validate a nonlinear control approach for 3-D bipedal walking that achieves asymptotic global-position tracking by explicitly addressing the hybrid robot dynamics.
The main contributions of this study are summarized as follows:
\begin{enumerate}
\item[i)] 
    Constructing impact invariance conditions that are independent of the desired global-position trajectory and yet ensure {\it all} desired trajectories respect the impact dynamics. 
    They can be used to decouple the planning of virtual constraints and global position, thus improving trajectory generation efficiency.
	\item[ii)] Establishing sufficient conditions based on the multiple Lyapunov stability analysis~\cite{branicky1998multiple} of the hybrid system for guiding the design of a continuous state-feedback control law to achieve asymptotic global-position tracking. 
	\item[iii)] Demonstrating the global-position tracking accuracy of the proposed control approach both through simulations and, for the first time, experimentally on a 3-D bipedal walking robot. 
	\item[iv)] Experimentally validating the inherent robustness of the proposed control design in addressing irregular walking surfaces such as moderately slippery floors.
\end{enumerate}

Some of the results presented in this paper were initially reported in \cite{gu2018straight} and \cite{gao2019time}. 
The present paper includes substantial, new contributions in the following aspects:
a) the proof of the main theorem (i.e., Theorem~\ref{theorem2}) is updated with a new choice of Lyapunov function to properly analyze the convergence of the robot's lateral foot placement during 3-D walking, 
and Proposition~\ref{prop-y-bound} is added along with its full proof, which supports the updated proof of the main theorem;
b) fully developed proofs of all theorems and propositions are presented, which were missing in \cite{gu2018straight} and \cite{gao2019time};
c) comparative experimental results are added to show the reliable global-position tracking performance of the proposed control approach;
and d) robustness evaluation is newly included to illustrate the capability of the proposed control approach in handling relatively slippery grounds.

This paper is structured as follows.
Section~\ref{sec-problem} describes the problem formulation.
Section~\ref{sec-control} explains the proposed continuous-phase tracking control law.
Section~\ref{sec-impact} presents the proposed construction of conditional impact invariance for designing virtual constraints.
Section~\ref{sec-stability} introduces the closed-loop stability analysis based on multiple Lyapunov functions.
Section~\ref{sec-sim} reports the simulation and experimental results. 
{Section~\ref{sec-discussions} discusses {the proposed approach and} potential directions of future work}.
Proofs of all theorems and propositions are given in the appendix.

\section{Problem Formulation}
\label{sec-problem}

This section presents the proposed problem formulation of global-position tracking control, including dynamics modeling, tracking error definition, and control objective.

\subsection{Full-Order Robot Model}

This subsection describes a full-order model that accurately captures the dynamic behaviors of all degrees of freedom (DOFs) involved in bipedal walking.
Thanks to the model's accuracy, a controller that is effective for the model would also be valid for the physical robot.
Hence, we use the full-order model as a basis of the proposed control approach.

The full-order model is naturally hybrid and nonlinear, because walking dynamics are inherently hybrid, involving both nonlinear continuous behaviors (e.g., leg-swinging motions) and state-triggered discrete behaviors (e.g., the joint-velocity jumps caused by foot-landing impact).

In this study, we assume that the swing and the stance leg immediately switch roles upon a foot landing, with the new swing leg beginning to move in the air and the new stance leg remaining in a full, static contact with the ground until the next landing occurs~\cite{westervelt2003hybrid}.
The assumption is valid when the double-support phase is sufficiently short and when the stance foot does not notably slip on the ground.

Under this assumption, if all of the robot's (revolute or prismatic) joints are directly actuated,
then the robot is fully actuated; i.e., its full DOFs can be directly commanded within continuous phases.

This study focuses on the relatively simple gait, fully actuated gait, for two main reasons.
First, asymptotic tracking of time-varying global-position trajectories {for the 3-D hybrid walking model} is still an open control problem for this simple gait.
Second, using a simple gait allows us to focus on addressing the complexity of the controller design problem induced by the hybrid, nonlinear robot dynamics and the time-varying global-position trajectory.

\begin{figure}[t]
	\centering
	\includegraphics[width=0.6\linewidth]{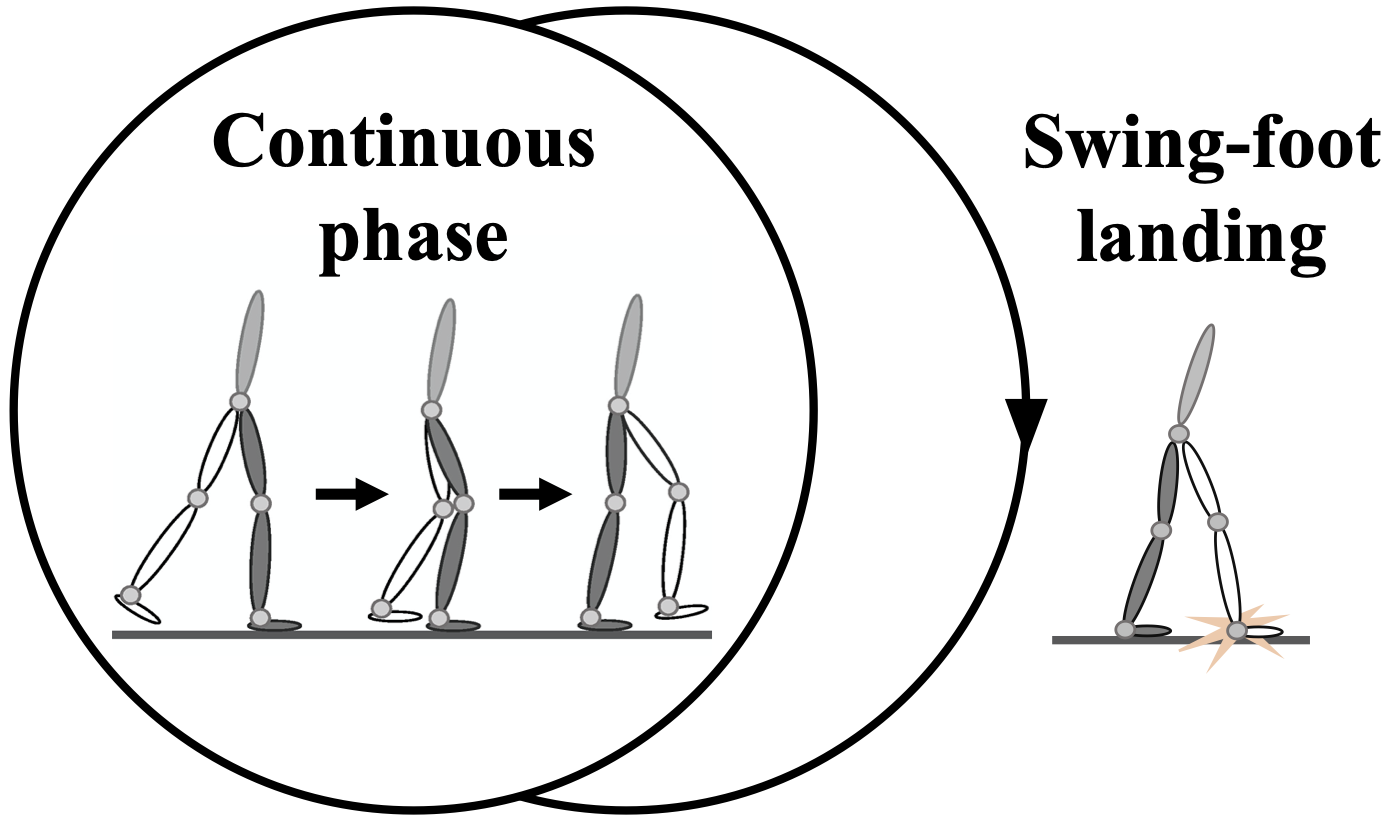}
	\caption{
		Illustration of a fully actuated gait cycle comprising a continuous phase and a discrete swing-foot landing.
		}
	\label{fig-FA}
\end{figure}

\noindent \textbf{Continuous-phase dynamics.} 
As illustrated in Fig.~\ref{fig-FA}, a complete walking cycle comprises:
a) a fully-actuated continuous phase during which one foot contacts the ground and the other swings in the air and
b) a landing impact.

Walking dynamics during continuous phases can be described by usual ordinary differential equations.
Lagrange's method is used to obtain the {following nonlinear full-order} model during continuous phases~\cite{grizzle2001asymptotically}:
\begin{equation}
	\bM(\bq) \ddot{\bq} + \bc(\bq,\dot{\bq}) =\bB_u \bu,
	\label{c5-continuous}
\end{equation}
where 
$\bq \in Q$ is the joint-position vector,
$\bM: Q \rightarrow   \mathbb{R}^{n\times n}$ is the symmetric, positive-definite inertia matrix,
$\bc : TQ \rightarrow   \mathbb{R}^n$ is the sum of Coriolis, centrifugal, and gravitational terms,
$\bB_u \in \mathbb{R}^{n\times m} $ is the joint-torque projection matrix {with full column rank},
and
$\bu \in U$ is the joint-torque vector.
Here,
$Q \subset  \mathbb{R}^n$ is the configuration space of the robot,
$TQ$ is the tangent {bundle} of $Q$,
and $U  \subset   \mathbb{R}^m$ is the admissible joint-torque set.
Note that $m=n$ when a robot is fully actuated.

\noindent \textbf{Impact dynamics.} When the swing foot lands on the ground, the swing and stance legs immediately switch their roles.
Here we model the swing-foot landing impact as the contact between rigid bodies~\cite{westervelt2003hybrid}.
This assumption is valid for dynamic walking on relatively stiff surfaces (e.g., concrete and ceramic floors) during which the swing foot strikes the surface at a relatively significant downward velocity.

Due to the coordinate swap of the swing and stance legs as well as the impulsive rigid-body impact, both joint position and velocity vectors experience a sudden jump at a landing event.
This state-triggered jump is described by the following nonlinear reset map $\bm \Delta_{q,\dq}: TQ \rightarrow TQ$~\cite{grizzle2001asymptotically}:
\begin{equation}
	\begin{bmatrix}
		\bq^+ \\
		\dot{\bq}^+
	\end{bmatrix}
	=
	\bm \Delta_{q,\dq} (\bq^-,\dot{\bq}^-)
	:=
	\begin{bmatrix}
		\bm \Delta_{q} (\bq^-) \\
		\bm \Delta_{\dq} (\bq^-)\dot{\bq}^-
	\end{bmatrix}
	,
	\label{c5-reset}
\end{equation}
where $\star^-$ and $\star^+$ represent the values of $\star$ just before and just after the impact, respectively.

\noindent \textbf{Switching surface.} A swing-foot landing event is triggered when the robot's state reaches the switching surface $S_q$, which is given by:
\begin{equation}
	S_q:=
	\lbrace ( \bq, \dot{\bq} ) \in TQ: z_{sw}(\bq)=0,~
	\dot{z}_{sw}  ( \bq , \dot{\bq}) <0  \rbrace,
	\label{c5-eq2}
	\vspace{+0.05in}
\end{equation}
where $z_{sw}:Q \rightarrow \mathbb{R}$ is the swing-foot height above the ground.

Combining the above equations yields the following full-order model:
\begin{equation}
	\begin{aligned} 
		\begin{cases}
			\bM(\bq) \ddot{\bq} + \bc(\bq,\dot{\bq}) =\bB_u \bu, 
			&\mbox{if } (\bq^-, \dot{\bq}^-) \notin S_q;
			\\
			\begin{bmatrix}
				\bq^+ \\
				\dot{\bq}^+
			\end{bmatrix}
			=
			\bm \Delta_{q,\dq} (\bq^-, \dot{\bq}^-), &\mbox{if } (\bq^-, \dot{\bq}^-) \in S_q.
		\end{cases}
	\end{aligned}
	\label{c5-eq1}
\end{equation}

\subsection{Global-Position Tracking Error}

A fully-actuated, $n$-DOF bipedal robot can track $n$ independent desired position trajectories, including the reference global-position trajectories.

In this study, we choose to use the position of a biped's {base} (e.g., trunk), $(x_b,y_b,z_b)$, 
to represent its global position in an environment.
The horizontal components of the {base} position are related to the stance-foot position as:
\begin{equation}
	x_b = x_{st} + \bar{x}_b (\bq)~\text{and}~
	y_b =y_{st} + \bar{y}_b (\bq),
\end{equation}
where 
$(x_{st},y_{st},0)$ denotes the stance-foot position with $x_{st},y_{st} \in \mathbb{R}$.
The scalar variables
$ \bar{x}_b  : Q \rightarrow Q_x \subset \mathbb{R}$ and $ \bar{y}_b: Q \rightarrow Q_y \subset \mathbb{R}$ represent the $x$- and $y$-coordinates of the base position relative to the stance foot, respectively.

In real-world locomotion tasks, a higher-level planner typically specifies the desired global motions as:
\begin{enumerate}
	\item[a)] The center line $\Gamma_d$ of the desired global path.
	\item[b)] The desired smooth position trajectory $s_{d}(t)$ along $\Gamma_d$.
\end{enumerate}
{As an arbitrary curved path can be approximated as a nonsmooth curve pieced together by straight lines,
this study focuses on the tracking control of straight-line paths}, 
{which could be extended to the tracking of a curved path as discussed in Section~\ref{sec-discussions}.}

Without loss of generality, suppose that 
the center line $\Gamma_d$ coincides with the $X_w$-axis of the world frame; that is
$$
\Gamma_d = \{(x_b,y_b) \in \mathbb{R}^2:  y_b =  0  \}.
\label{Gamma}
$$
Then, the global-position tracking error along $\Gamma_d$ is defined as
$
\bar{x}_b (\bq)-(s_{d}(t) - x_{st} )
$.

While $s_d(t)$ and $\Gamma_d$ are often provided by a higher-level path planner, the desired base motion in the direction lateral to $\Gamma_d$ remains to be designed, which is explained next.

\subsection{Virtual-Constraint Tracking Error}

Besides the desired global-position trajectory $s_d(t)$, a legged robot typically has multiple directly actuated DOFs that can track additional desired motions.
We choose to use virtual constraints to define the desired trajectories for the lateral base position $y_b$ and the remaining control variables $\bm \phi_{c} : Q \rightarrow Q_{c} \subset \mathbb{R}^{n-2}$.

Analogous to the HZD framework, we use virtual constraints to represent the desired configuration relative to a phase variable $\theta:Q \rightarrow Q_f \subset \mathbb{R}$.
Without loss of generality, the phase variable $\theta$ is chosen as the relative forward position of the base, $\bar{x}_b (\bq)$; that is, 
$$\theta = \bar{x}_b (\bq).$$
The virtual constraints can be encoded by $\theta$ as:
\begin{equation}
	\begin{bmatrix}
		\bar{y}_b (\bq) \\
		\bm \phi_{c}(\bq)
	\end{bmatrix}
	-
	\begin{bmatrix}
		{y_d( \theta (\bq) )}- y_{st} \\
		\bm \phi_d (\theta(\bq))
	\end{bmatrix}
	= \bzero,
	\label{c5-vir}
\end{equation}
\noindent where
the scalar function
$ y_d: Q_f \rightarrow \mathbb{R} $ and the vector $\bm \phi_d : Q_f \rightarrow \mathbb{R}^{n-2}$
are the desired trajectories of $y_b$ and $\bm \phi_{c}$, respectively.
Suppose that $y_d$, $\bm \phi_d$, $\bm \phi_{c}$ and $\theta$ are all continuously differentiable in their respective arguments.

Thus, the tracking error corresponding to the virtual constraints is defined as
$
\begin{bmatrix}
	\bar{y}_b (\bq) \\
	\bm \phi_{c}(\bq)
\end{bmatrix}
-
\begin{bmatrix}
	{y_d( \theta (\bq) )}- y_{st} \\
	\bm \phi_d (\theta(\bq))
\end{bmatrix}
$.

\subsection{Control Objective}

The tracking errors can be compactly expressed as:
\begin{equation}
	\bh(t,\bq) =
	\bh_c(\bq) - \bh_d(t,\bq),
	\label{output fcn}
\end{equation}
where the control variables $\bh_c$ and their desired trajectories $\bh_d$ are respectively defined as
$
\bh_c:= \begin{bmatrix}
	\bar{x}_b,
	\bar{y}_b,
	\bm \phi_{c}^T
\end{bmatrix} ^T
$
and 
$
\bh_d:=\begin{bmatrix}
	s_{d} - x_{st},
	y_d- y_{st},
	\bm \phi_d^T
\end{bmatrix}^T
$.

The control objective is to asymptotically drive the tracking error $\bh$ to zero
for achieving asymptotic tracking of the desired motions, which are the desired global-position trajectory $s_d(t)$ and the desired functions $y_d$ and $\boldsymbol{\phi}_d$ that define the virtual constraints, for 3-D bipedal robot walking.

\textcolor{black}{To achieve this objective, the proposed control approach (Fig.~\ref{overview}) comprises three main components: a) continuous-phase controller design ({for stabilizing the desired trajectories within continuous phases}); b) impact invariance construction ({for satisfying a necessary condition of asymptotic tracking for the hybrid model}); and c) closed-loop stability analysis ({for providing sufficient stability conditions that guide the controller design}).}

\section{Continuous-Phase Control}
\label{sec-control}
This section presents a continuous state-feedback control law {that asymptotically stabilizes the desired trajectories within {\it continuous} phases}.

We choose to design a controller that directly regulates the continuous-phase walking dynamics instead of the impact dynamics because the impact dynamics cannot be directly commanded due to its infinitesimal short duration.
We will show in Section~\ref{sec-stability} how the proposed continuous control law could be tuned to indirectly stabilize the desired trajectories for the overall hybrid system.

\vspace{0in}
\begin{figure}[t]
	\centering
	\includegraphics[width=1\linewidth]{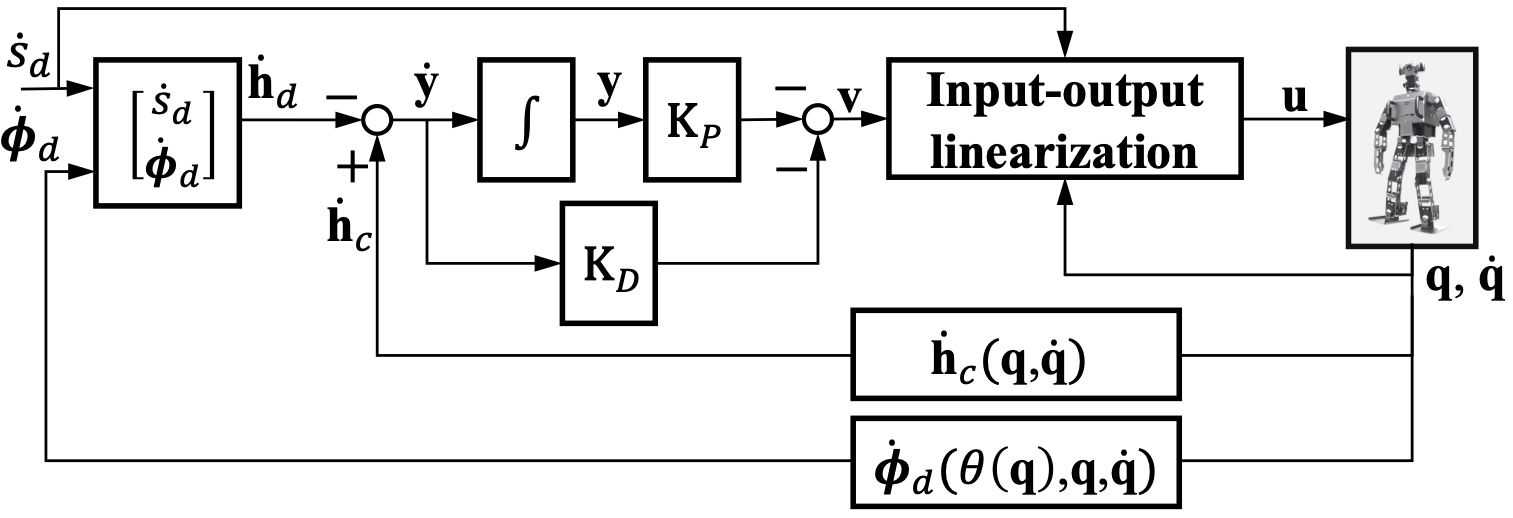}
	\caption{
		{Block diagram of the proposed continuous-phase controller.}
	}
	\label{ControlScheme}
\end{figure}
\vspace{0in}

The proposed control law ({Fig.}~\ref{ControlScheme}) is synthesized based on the full-order model of bipedal walking dynamics.
Analogous to the HZD framework, 
we utilize the
input-output linearization technique~\cite{khalil1996nonlinear} to linearize the nonlinear continuous-phase dynamics in Eq.~\eqref{c5-continuous} into a linear map,
which allows us to exploit the well-studied linear system theory to design the needed controller for the continuous phase.

With the trajectory tracking error $\bh$ chosen as the output function $\by$ (i.e., $\by=\bh$),
a continuous-phase control law synthesized via input-output linearization is given by:
\begin{equation}
	\begin{aligned}
	\bu = ( \bJ_h \bM^{-1} \bB )^{-1}
		(\bv + \tfrac{\partial \bh }{\partial \bq } \bM^{-1} \bc - \tfrac{\partial^2 \bh  }{\partial t^2}
		-\tfrac{\partial}{\partial \bq} (\tfrac{\partial \bh}{\partial \bq} \dot{\bq} ) \dot{\bq} )
		\label{control-law}
	\end{aligned}
\end{equation}
with
$
		\bJ_{h} (\bq):= 
		\frac{ \partial {\bh} }{ \partial \bq  } (\bq),
$
which yields the linearized dynamics $\ddot{\by} = \bv$.
Note that the variables $\bm \phi_c$, $y_d$, and $\bm \phi_d$ can be chosen such that there exists an open subset $\tilde{Q}$ of the configuration space $Q$ on which the Jacobian matrix $\bJ_{h} (\bq)$ is invertible.
Then, the matrix $\bJ_h \bM^{-1} \bB$ is invertible on $\bq \in \tilde{Q}$.


Choosing $\bv$ as a proportional-derivative (PD) term
\begin{equation}
	\bv = -\bK_P \by - \bK_D \dot{\by},
	\label{c5-v}
\end{equation}
where the proportional gain matrix $\bK_P \in \mathbb{R}^{n \times n}$ and the derivative gain matrix $\bK_D \in \mathbb{R}^{n \times n}$ are both positive-definite diagonal matrices,
the linear closed-loop dynamics of the output function becomes $\ddot{\by} + \bK_D \dot{\by} + \bK_P \by = \bzero$ during continuous phases.

Define the state of the output function dynamics as
$\bx := \begin{bmatrix}
	\by^T,
	\dot{\by}^T
\end{bmatrix}^T
\in \mathbb{R}^{2n}$.
Then, the closed-loop error equation can be compactly expressed as: 
\begin{equation}
	\begin{aligned}
		\begin{cases}
			\dot{\bx} = \bA \bx,~&\mbox{if}~(t^-,\bx^-) \notin S; \\
			\bx^+ = \bm \Delta (t^-,\bx^{-}),~&\mbox{if}~(t^-,\bx^{-}) \in S.
		\end{cases}
	\end{aligned}
	\label{c5-hybrid}
\end{equation}
Here,
$
\bA
:=
\begin{bmatrix}
	\bzero &\bI \\
	-\bK_P &- \bK_D
\end{bmatrix}
$
with $\bzero$ a zero matrix and $\bI$ an identity matrix with appropriate dimensions.
$S$ and $\bm \Delta$ are the switching surface and impact map associated with the closed-loop dynamics, respectively.
Note that $\bm \Delta$ is explicitly time-dependent because of the explicit time dependence of $\bh$. 
The expressions of $S$ and $\bm \Delta$ can be obtained from their counterparts in the open-loop dynamics (Eq.~\eqref{c5-eq2}) as well as the output function definition (Eq.~\eqref{output fcn}).

The origin (i.e., $\bx=\bzero$) of the continuous-phase closed-loop dynamics (i.e., $\dot{\bx} = \bA \bx$) will be asymptotically stable if the PD gains are chosen such that $\bA$ is Hurwitz~\cite{khalil1996nonlinear}.
Then, there exist positive numbers $c_{1}$, $c_{2}$, and $c_{3}$ and a Lyapunov function candidate $V(\bx)$ such that
\begin{equation}
	c_{1} \|\bx\|^2 \leq V(\bx) \leq c_{2} \|\bx\|^2~\mbox{and}~\dot{V}(\bx) \leq - c_{3}  \|\bx\|^2
	\label{c5-eq17}
\end{equation}
hold for all $\bx$ within continuous phases.
These inequalities indicates that $V(\bx)$ exponentially converges at the rate of $\frac{c_3}{c_2}$ within a continuous phase.

While the proposed control law with properly chosen PD gains guarantees the asymptotic tracking of the desired trajectories within continuous phases, the impact dynamics (i.e., $\bx^+ = \bm \Delta(t,\bx^{-})$) remain uncontrolled, and thus the stability of the hybrid closed-loop system is not yet ensured. 
To satisfy a necessary condition of asymptotic trajectory stabilization in the presence of uncontrolled impact dynamics, we introduce impact invariance construction next.

\section{Impact Invariance Construction for Virtual Constraint Design}
\label{sec-impact}

This section presents the proposed construction of impact invariance conditions
that can be incorporated in the trajectory generation of the desired functions $y_d$ and $\boldsymbol{\phi}_d$, which define the virtual constraints, for ensuring all desired trajectories (i.e., $y_d$ and $\boldsymbol{\phi}_d$ as well as $s_d$) respect the impact dynamics.

\subsection{Impact Invariance}
The concept of impact invariance was first introduced within the HZD framework, along with a systematic method of impact invariance construction (see Theorem 4 in~\cite{westervelt2003hybrid}).
The concept was later on termed as ``impact invariance''~\cite{morris2009hybrid}.

\begin{defn}
\textbf{(Impact invariance condition)}
	The output function $\by$ and its first derivative $\dot{\by}$ are {\it impact-invariant} if the following condition is met: $\by^+=\bzero$ and $\dot{\by}^+=\bzero$ hold just after an impact if $\by^-=\bzero$ and $\dot{\by}^-=\bzero$ hold just before an impact~\cite{westervelt2003hybrid,morris2009hybrid}.
\end{defn}

For the proposed feedback controller to achieve asymptotic tracking for hybrid dynamical systems, the output function state $\by$ and $\dot{\by}$ need to satisfy the impact invariance condition at the steady state.
Suppose that the impact invariance condition is not met at the steady state.
Then, because the robot's impact dynamics cannot be directly regulated, the output function state may become nonzero just after an impact even if it is zero just before the impact, which means asymptotic tracking cannot be achieved.

Since the impact invariance condition is placed on the output function state,
we can satisfy it by properly planning the desired function $\bh_d$.
As the desired global position $s_d$ is often supplied by a higher-level path planner without impact dynamics considered, 
the generation of the remaining desired functions $y_d$ and $\bm \phi_d$, which define the virtual constraints, needs to ensure the impact agreement for all trajectories (i.e., $s_d$, $y_d$, and $\bm \phi_d$).

The proposed impact invariance construction comprises two steps.
We first extend the existing method (i.e., Theorem 4 in~\cite{westervelt2003hybrid}) to derive conditions that ensure the impact invariance of the output function state associated with the virtual constraints,
that is,
($\bar{y}_b -(y_d- y_{st})$, $\bm \phi_{c}-\bm \phi_d$)
and its first derivative (Section~\ref{sec-impact-virtual}).
Built upon this condition, we introduce a new, additional condition that guarantees the impact invariance of the global-position error state, that is, $\bar{x}_b -(s_{d} - x_{st} )$ and its first derivative (Section~\ref{sec-impact-global}).
Both conditions are placed on the virtual constraints alone.

\subsection{Impact Timings}
\label{sec-impact-timing}

Because the desired global-position trajectory $s_d$ is explicitly time-varying, we need to consider the impact timings in the proposed impact invariance construction.
As the actual and desired impact timings generally do not coincide due to the state-triggered nature of a foot-landing event~\cite{rijnen2019hybrid}, they are individually defined as follows.
\vspace{0in}
\begin{figure}[t]
	\centering
	\includegraphics[width=0.9\linewidth]{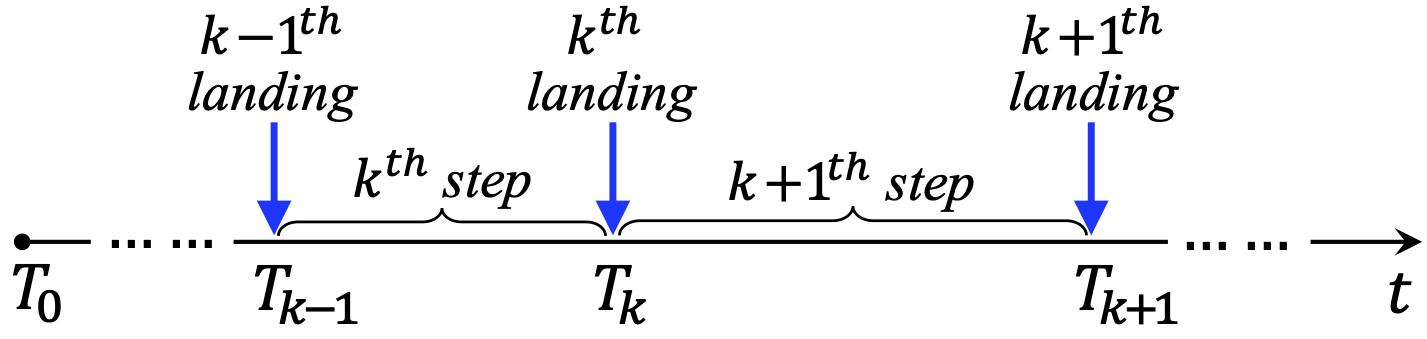}
	\caption{
		Illustration of the impact timings of actual walking steps.
		The actual $k^{th}$ walking step begins at $t=T_{k-1}^+$ and ends at $t=T_{k}^-$.
		The actual $k^{th}$ swing-foot landing occurs at $t=T_{k}^-$.  
	}
	\label{fig-Tk}
\end{figure}
\vspace{0in}

\begin{defn}
(\textbf{Actual and desired impact timings})
	Let $T_k$ be the timing of the $k^{th}$ ($k \in \mathbb{Z}^+$) actual landing impact, which is defined as the timing of the first intersection between the state $\bx$ and the switching surface $S$ on $t > T_{k-1}^+$.
	Without loss of generality, define $T_0=0$.
	Let $\tau_k$ denote the $k^{th}$ desired impact timing, which is defined as the timing of the first intersection between $\bx$ and $S$ on $t > T_{k-1}^+$ assuming $\bx=\bzero$ $\forall t > T_{k-1}^+$. 
\end{defn}

The precise definition of $T_k$ is given in~\cite{westervelt2003hybrid}.
Figure~\ref{fig-Tk} shows an illustration of $T_k$.
The variables $\star(T_{k-1}^-)$ and $\star(T_{k-1}^+)$ are respectively denoted as $\star|^-_{k-1}$ and $\star|^+_{k-1}$ in the rest of the paper where brevity is preferred.

\subsection{Impact Invariance for Virtual Constraints }
\label{sec-impact-virtual}

The proposed impact invariance construction utilizes the uniqueness of the robot's joint position $\bq^*$ just before an impact event when the virtual constraints in Eq.~\eqref{c5-vir}
are exactly satisfied~\cite{westervelt2003hybrid}.

{The joint position $\bq^*$ is mathematically defined as the solution to the following equations}
\begin{equation}
\bF(\bq):=
	\begin{bmatrix}
		\bar{y}_b (\bq) - ( {y_d( \theta (\bq) )}- y_{st} ) \\
		\bm \phi_{c}(\bq) - \bm \phi_d (\theta(\bq)) \\
		z_{sw}(\bq)
	\end{bmatrix}
	= \bzero
	\label{c5-vir}
\end{equation}
on $S \cap \tilde{Q}$.
Note that the last equation in Eq.~\eqref{c5-vir} holds because the swing-foot height $z_{sw}(\bq)$ reaches 0 at a touchdown.

Due to the nonlinearity of the function $\bF(\bq)$, Eq.~\eqref{c5-vir} may have multiple solutions on $S \cap \tilde{Q}$.
Suppose that the output function is designed such that $\frac{\partial \bF}{\partial \bq}(\bq^*)$ is invertible on $S \cap \tilde{Q}$.
Then by the implicit function theorem, there exits $\bar{Q} \subset \tilde{Q}$ such that $\bq^*$ is a unique solution to $\bF(\bq)=\bzero$ on $S \cap \bar{Q}$.

We are now ready to introduce the condition that ensures the impact invariance of the output function state associated with the virtual constraints.

\begin{prop}
\textbf{(Impact invariance conditions for virtual constraints)}
\label{prop2}
	Suppose that the desired functions $y_d$ and $\bm \phi_d$ are planned to meet
	the following conditions:
	{
		\begin{enumerate}
			\item [(A1)]  
			$ \bar{y}_b ( \bq_0    )  =   y_d ( \theta_0 ) - y_{std}$ and $	\bm \phi_c( \bq_0    )  =   \bm \phi_d ( \theta_0 ) $.
		\item[(A{2})] 
$
\begin{aligned}
&
\begin{bmatrix}
\frac{\partial \bar{y}_b}{\partial \bq}	\bar{y}_b (\bq_0) 
\\
\frac{\partial \bm \phi_{c}}{\partial \bq}	(\bq_0)
\end{bmatrix}
\bm \Delta_{\dq} (\bq^*)  
			\bJ_h^{-1} ( \bq^*)
			\begin{bmatrix}
			\small
				1 \\
				\frac{ \partial y_d  }{ \partial \theta} ( \theta^* ) \\
				\frac{ \partial \bm \phi_d   }{ \partial \theta}( \theta^* ) \\
			\end{bmatrix} 
\\
&=
\begin{bmatrix}
\small
\frac{ \partial y_d  }{ \partial \theta} ( \theta_0  )  
\\
\frac{ \partial \bm \phi _d  }{ \partial \theta} ( \theta_0)
\end{bmatrix}
\frac{ \partial \bar{x}_b }{ \partial \bq  }  (    \bq_0   ).
\end{aligned}
$
\end{enumerate}}
Here, $ \bq_0 :=  \bm \Delta_{q} (\bq^*)$,
$\theta_0 := \bar{x}_b(\bq_0)$,
and
$\theta^*:=\theta(\bq^*)$.
The scalar $y_{std}$ is the $y$-coordinate of the desired foot placement.
Then, under the lateral foot-placement condition $y_{st} = y_{std}$, impact invariance holds for the output function state associated with the virtual constraints.
	\label{theorem 1}
\end{prop}

\subsection{Impact Invariance for Global-Position Tracking Error}
\label{sec-impact-global}

As the desired global-position trajectory $s_d$ is often supplied by a high-level planner without impact dynamics considered, 
we construct an additional condition, which is placed on the virtual constraints,
to ensure the impact invariance of the global-position error state, i.e., $\bar{x}_b -(s_{d} - x_{st} )$ and its first derivative.
Note that $\bar{x}_b -(s_{d} - x_{st} ) \equiv {x}_b - s_d $.

The key to the proposed construction is to exploit the property of $s_d$ that it is commonly planned as a smooth function for any $t>T_0$.
Thanks to this property, the impact invariance of the output function ${x}_b - s_d $ is always guaranteed; that is, ${x}_b - s_d =0$ automatically holds just after an impact if it holds just before the impact.
This is because both the forward base position ${x}_b$ and its desired trajectory $s_d$ are continuous across an impact event.

We choose to ensure the impact invariance of $\dot{{x}}_b - \dot{s}_d $ by enforcing the continuity of $\dot{{x}}_b$ across the planned impact event.
The rationale of this design choice is threefold.
First, given the continuity of $\dot{s}_d$ for any $t>T_0$, the continuity of $\dot{x}_b$ across the planned impact event guarantees the continuity of $\dot{{x}}_b - \dot{s}_d $, which then ensures that $\dot{x}_b - \dot{s}_d =0$ holds just after the planned impact if it holds just before the impact.
Second, the continuity of $\dot{x}_b$ is equivalent to that of $\dot{\bar{x}}_b$ because the stance foot does not move (i.e., $\dot{x}_{st}=0$).
Third, $\dot{\bar{x}}_b$ is a function of the joint position $\bq$ and velocity $\dot{\bq}$ only, and thus its continuity across the planned impact event can be satisfied through virtual constraint design alone without explicitly relying on the profile of $s_d$.

The proposed impact invariance condition for $\dot{{x}}_b - \dot{s}_d $ is summarized as follows.

\begin{prop}
(\textbf{Impact invariance condition for global-position error})
Suppose that the desired functions $y_d$ and $\bm \phi_d$ satisfy conditions (A1) and (A2) and the following condition:
\begin{enumerate}
			\item [(A3)]  
			$
			\frac{ d \bar{x}_b }{ d \bq  }  (   \bq_0    )      \bm \Delta_{\dq} (\bq^*)  \bJ_h^{-1} ( \bq^* )
			\begin{bmatrix}
				1
				\\
				\frac{ d y_d  }{ d \theta} ( \theta^* ) 
				\\
				\frac{ d \bm \phi_d  }{ d \theta} ( \theta^* ) 
			\end{bmatrix}  
			= 1.
			$
\end{enumerate}
Then, under the lateral foot-placement condition $y_{st} = y_{std}$, impact invariance of the global-position error state holds.
\label{theorem1-b}
\end{prop}

If the virtual constraints are generated to meet the conditions in Propositions~\ref{theorem 1} and~\ref{theorem1-b}, then under the lateral foot-placement condition $y_{st} = y_{std}$, the impact invariance of the full output function state holds;
that is, if $\bx(\tau_k^-)=\bzero$ then $\bx(\tau_k^+)=\bm \Delta(\tau_k^-,\bzero) = \bzero$.

\begin{rem} (\textbf{Independence from desired global-position trajectory})
Propositions~\ref{theorem 1} and~\ref{theorem1-b} indicate that the satisfaction of the impact invariance conditions only relies on the design of the virtual constraints but not the arbitrary global-position trajectory $s_d$ provided by a higher-level planner.
For this reason, the design of virtual constraints does not need to explicitly consider $s_d$ and thus can be performed offline even when the higher-level planner updates $s_d$ online, which could reduce the computational load for online planning.
\end{rem}

\begin{rem}
(\textbf{Ensuring the desired lateral foot placement through controller design})
Note that the foot-placement condition $y_{st} = y_{std}$ underlying the proposed impact invariance construction is only assumed in the virtual constraint planning but not the controller design.
Indeed, Section~\ref{sec-stability} introduces sufficient conditions under which the proposed controller guarantees this foot-placement condition holds at the actual steady state.
\end{rem}

\section{Stability Analysis }

\label{sec-stability}

This section introduces Lyapunov-based stability analysis of the hybrid, nonlinear, time-varying closed-loop error dynamics (Eq.~\eqref{c5-hybrid}) under the proposed continuous-phase control law (Eqs.~\eqref{control-law} and \eqref{c5-v}).
The outcome of this stability analysis is a set of sufficient conditions under which the proposed control law provably realizes asymptotic stabilization of the desired global position trajectory $s_d$ and the desired functions $y_d$ and $\bm \phi_d$ for the overall hybrid system.

\subsection{Boundedness of Foot Placement and Impact Timing}

Before presenting the main theorem on closed-loop stability, we first introduce the boundedness of {the impact timing $T_k$} and the lateral stance-foot position $y_{st}$.
The boundedness of the impact timing is needed in the stability analysis to derive how much a Lyapunov function converges within a continuous phase.
The boundedness of $y_{st}$ also needs to be explicitly considered,
because $y_{st} = y_{std}$ underlies the proposed impact invariance conditions and should hold at the actual steady state for achieving asymptotic tracking.

\begin{prop}\textbf{(Boundedness of impact timing error)}
\label{prop-t-bound}
Let
$\tilde{\bx}(t;t_0,\lambda_0)$
be a solution of a fictitious continuous-time system $\dot{\tilde{\bx}} = \bA \tilde{\bx}$ with the initial condition $\tilde{\bx}(t_0)=\lambda_0$, $\forall t>t_0$.
There exists {a positive number $r_1$ and a Lipschitz constant $L_{T_x}$} such that the difference between the actual and planned impact timings is bounded above in norm as
\begin{equation}
		| T_k - \tau_k | \leq L_{T_x} \|  \tilde{\bx}(\tau_k;  T_{k-1}^+,\bx|^+_{k-1} )   \| 
		\label{c5-T}
\end{equation}
for any $\bx|_{0}^+ \in B_{r_1} (\bzero) := \{  \bx \in \mathbb{R}^{2n} :  \| \bx \| \leq r_1    \}$ and any $k \in \mathbb{Z}^+$.
\end{prop}

\begin{prop} \textbf{(Boundedness of lateral foot-placement error)}
	Suppose that the lateral swing-foot position $y_{sw}$ is chosen as an element of $\bm \phi_c$ and is thus directly controlled.
	Then, there exist positive numbers $\beta_{st}$ and $d_1$ such that
	{the foot-placement error after the $k^{th}$ swing-foot landing} is bounded above in norm as
	\begin{equation}
		| y_{st}|_{k}^+ - y_{std} | \leq  \| \bx|^-_{k} \| + \beta_{st}  \|  \tilde{\bx}(\tau_k;T_{k-1}^+, \bx|^+_{k-1}) \| 
		\label{yst-bound}
	\end{equation}
	for any
	$\bx|_{0}^+ \in B_{d_1} (\bzero) := \{  \bx \in \mathbb{R}^{2n} :  \| \bx \| \leq d_1    \}$ and any $k \in \mathbb{Z}^+$. 
	\label{prop-y-bound}
\end{prop}

\noindent \textbf{Rationale of proofs.}
The full proofs of Propositions~\ref{prop-t-bound} and \ref{prop-y-bound} are given in the appendix. 
The proof of Proposition~\ref{prop-t-bound} utilizes the implicit dependence of the actual impact timing $T_k$ on the error state $\bx$.
The proof of Proposition~\ref{prop-y-bound} mainly relies on the fact that the stance-foot position within the current step is the end position of the swing foot within the previous step.
By including $y_{sw}$ as a control variable, we can then relate the lateral foot-placement error $y_{st}-y_{std}$ to the error state $\bx$.
$\hfill \blacksquare$

\subsection{Main Theorem}

If the virtual constraints are designed to satisfy the impact invariance conditions in Propositions {\ref{theorem 1} and \ref{theorem1-b}} and if the continuous-phase convergence rate of $\bx$ is sufficiently fast, then the origin of the hybrid closed-loop error system is asymptotically stable, as summarized in the main theorem:

\begin{thm}(\textbf{Closed-loop stability conditions})
Suppose that the virtual constraints satisfy the impact invariance conditions (A1)-(A3). 
Also, suppose that the PD gains in Eq.~\eqref{c5-v} are chosen such that $\bA$ is Hurwitz and that the continuous-phase convergence rate of $\bx$ is sufficiently fast.
Then, there exists a positive number $d_2$ such that for any $\bx|_0^+ \in B_{d_2} (\bzero):=\{  \bx \in \mathbb{R}^{2n} :  \| \bx \| \leq d_2    \}$,
the {origin} of the closed-loop error system in Eq.~\eqref{c5-hybrid} is locally asymptotically stable; that is,  $\bx(t) \rightarrow \bzero ~ {as}~ t \rightarrow \infty.$
	
Furthermore, both the lateral foot placement and actual impact timing asymptotically converge to their desired values; that is,
$T_{k} - \tau_k \rightarrow 0 ~{and}~ y_{st} - y_{std}  \rightarrow 0  ~ {as}~ k \rightarrow \infty.$
	\label{theorem2}
\end{thm}

\noindent \textbf{Rationale of proof.}
The full proof of Theorem~\ref{theorem2} is given in the appendix. 
The proof utilizes the stability theory of the multiple Lyapunov functions~\cite{branicky1998multiple},
which prescribes how a Lyapunov function candidate should evolve in order for the origin of a hybrid dynamical system to be stable.

The stability analysis begins with the construction of the Lyapunov function candidate.
Since the lateral foot-placement error $y_{st} - y_{std}$ is not explicitly included in the state $\bx$ but directly affects the satisfaction of the impact invariance condition and thus the system stability,
we choose to construct the Lyapunov function $V_a$ by augmenting $V$ with
a positive-definite function of the foot-placement error:
	\begin{equation}
		V_a(\bx,  y_{st} - y_{std}):=V(\bx) + \sigma  (y_{st} - y_{std})^2,
		\label{Va}
	\end{equation}
where $\sigma$ is a positive number to be specified in the proof. 

Next, we analyze the evolution of $V_a$ during a continuous phase as well as through a hybrid transition.
The last step is to derive the sufficient closed-loop stability conditions that the continuous-phase convergence rate should meet such that the divergence of $V_a$ caused by the uncontrolled impact is compensated by the continuous-phase convergence.

The convergence of the foot placement $y_{st}$ and impact timing $T_k$ is proved based on Propositions~\ref{prop-t-bound} and~\ref{prop-y-bound} and the asymptotic convergence of the error state $\bx$.
By Propositions~\ref{prop-t-bound} and \ref{prop-y-bound}, the deviations of the lateral foot placement and impact timing are bounded above by the norms of the actual state $\bx$ and the fictitious state $\tilde{\bx}$.
Note that by definition, $\tilde{\bx}$ overlaps with $\bx$ within the given actual continuous phase.
Thus, driving $\bx$ to zero will indirectly make $\tilde{\bx}$ diminish, which then eliminates the deviations $y_{st}-y_{std}$ and $T_k-\tau_k$ at the actual steady state.
$\hfill \blacksquare$

\begin{rem} 
\label{rmk-PD}
(\textbf{Tuning continuous-phase convergence rate})
By Theorem~\ref{theorem2}, the continuous-phase convergence rate of $\bx$ (or equivalently, $V_a$) needs to be sufficiently fast for guaranteeing asymptotic trajectory tracking of the hybrid closed-loop system. 	
The continuous-phase convergence rate of $V_a$ solely depends on that of $V$,
because the stance foot is static during a continuous phase and $| y_{st}-y_{std} |$ remains constant. 
We can construct $V$ as $V=\bx^T \bP \bx$,
where $\bP$ is the solution to the Lyapunov equation~\cite{khalil1996nonlinear} $$\bP \bA + \bA^T \bP = -\bQ.$$
Here, $\bQ$ is any symmetric, positive-definite matrix satisfying $\bzero < \lambda_Q \bI \leq \bQ$ with a positive number $\lambda_Q$.
For simplicity, we can choose $\bQ$ as an identity matrix, and then $\lambda_Q$ can be any number satisfying $0 < \lambda_Q \leq 1$.
Then, the bounds of $V$ and $\dot{V}$ in Eq.~\eqref{c5-eq17} become $c_1=\lambda_{min}(\bP)$, $c_2=\lambda_{max}(\bP)$, and $c_3=\lambda_Q$, where $\lambda_{min}(\bP)$ and $\lambda_{max}(\bP)$ are the smallest and the largest eigenvalues of $\bP$, respectively.
Thus, the exponential convergence rate of $V$ becomes $\frac{c_3}{c_2}=\frac{\lambda_Q}{\lambda_{max}(\bP)}$.
Note that the value of the matrix $\bP$ depends on the PD gains, and thus $\lambda_{max}(\bP)$ can be adjusted by tuning those gains.
The full proof (Section~\ref{proof-thm2}) provides greater details about PD gain tuning.
It also provides an explicit expression of the lower bound of the convergence rate $\frac{c_3}{c_2}$ for guaranteeing asymptotic error convergence of the hybrid closed-loop system. 
\end{rem}

\begin{rem} (\textbf{Satisfying lateral foot-placement condition})
Theorem~\ref{theorem2} indicates that the lateral foot-placement condition underlying the proposed impact invariance construction in Propositions~\ref{theorem 1} and \ref{theorem1-b} is exactly met at the steady state.
Thus, the impact invariance of $\bx$, which is the necessary condition for asymptotic trajectory tracking, is indeed satisfied at the steady state; that is, if $\bx(\tau_k^-) \rightarrow \bzero$ then $\bx(\tau_k^+)=\bm \Delta(\tau_k,\bzero) \rightarrow \bzero$ as $k \rightarrow \infty$. 
\end{rem}

\section{Simulations and Experiments}
\label{sec-sim}

This section reports simulation and experimental results that demonstrate the global-position tracking performance of the proposed control approach.

The hardware platform used for controller validation is the OP3 bipedal humanoid robot developed by ROBOTIS ({Fig.~\ref{overview}}). 
OP3 weighs 3.5 kg, and its height is 0.51 m.
It has twenty revolute joints comprising eight upper-body and twelve leg joints.
Because OP3's twenty revolute joints are all independently actuated, the robot is fully actuated during a continuous phase.


\subsection{Virtual Constraint Generation}

This subsection explains the lower-level, optimization-based trajectory generation of virtual constraints based on the proposed impact invariance conditions.

With full actuation, OP3's twelve leg joints can be directly commanded to track twelve independent desired trajectories,
which are: 1) the desired global-position trajectory $s_d$ and 2) the desired functions $y_d$ and $\bm \phi_d$.
As a higher-level planner supplies the desired global path on the walking surface and the desired position trajectory along the path, the objective of the trajectory generation is to plan the desired lateral base position $y_d$ and desired function $\bm \phi_d$ that both define the virtual constraints.

\begin{figure}[t]
	\centering 
	\includegraphics[width=0.9\linewidth]{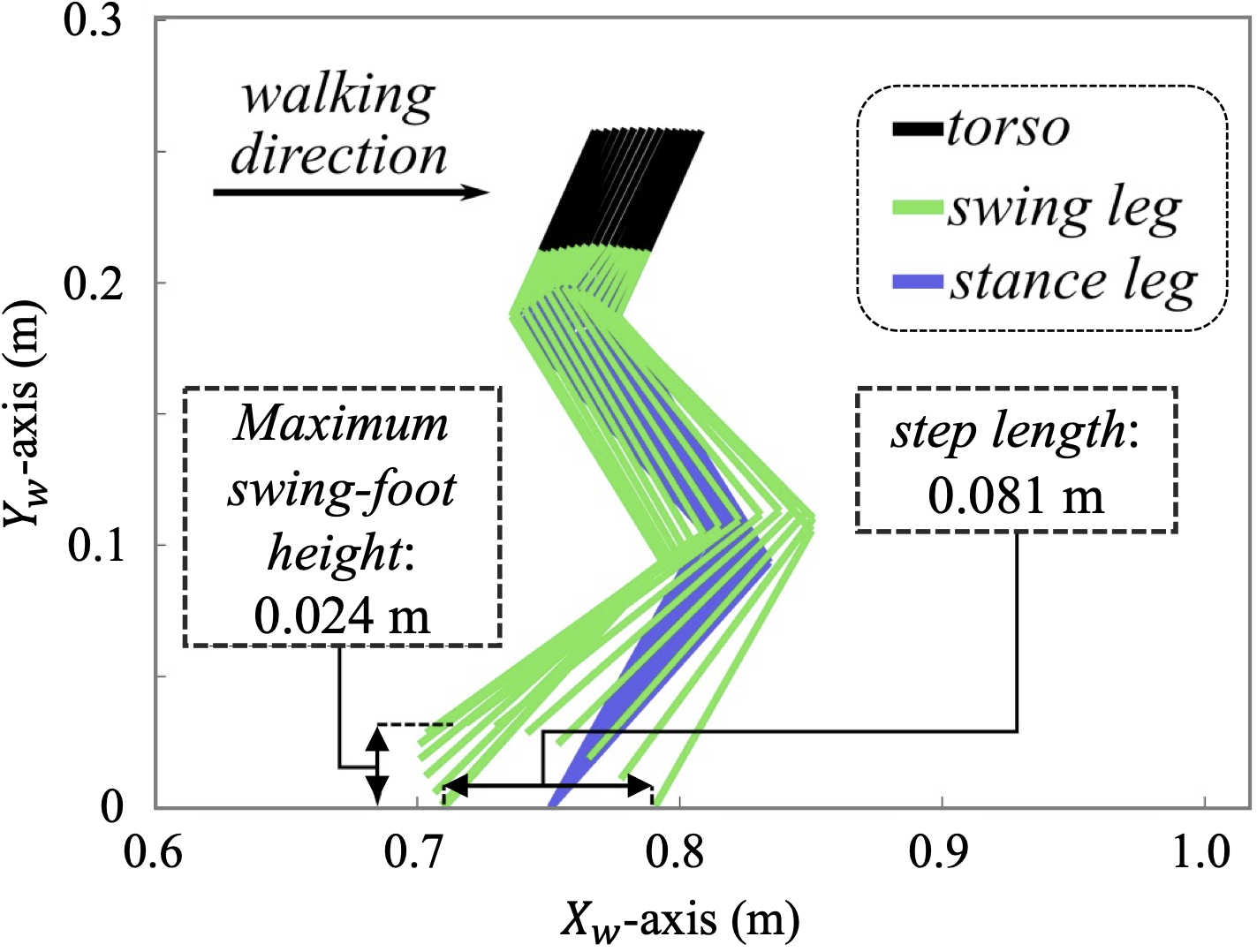}
	\caption{Illustration of the desired gait in the sagittal plane corresponding to the planned virtual constraints.}
	\label{fig4-pattern}
\end{figure}

\noindent \textbf{Trajectory parameterization.}
The desired lateral base position $y_d$ is chosen as the following simple sinusoidal function to enable an oscillatory global motion about the center line $\Gamma_d$ during 3-D walking:
\begin{equation}
	    y_d (\bar{x}_b):=	\alpha_{1} sin (\alpha_{2} \bar{x}_b + \alpha_{3}),
	\label{alpha}
	\end{equation}
where $\bm \alpha:=\begin{bmatrix}
		\alpha_1~\alpha_2~\alpha_3
	\end{bmatrix}^T \in \mathbb{R}^3$
is an unknown vector to be optimized.

The desired functions $\bm \phi_d$ are chosen as the desired trajectories for the following ten control variables $\bm \phi_c$: 
\begin{enumerate}
	\item[a)] Height ($z_b$) and roll, pitch, and yaw angles ($\psi_{b}^{roll}$, $\psi_{b}^{pitch}$, $\psi_{b}^{yaw}$) of the base.
	\item[b)] Position ($x_{sw}$, $y_{sw}$, $z_{sw}$) and roll, pitch, and yaw angles ($\psi_{sw}^{roll}$, $\psi_{sw}^{pitch}$, $\psi_{sw}^{yaw}$) of the swing foot.
\end{enumerate}
This choice of control variables allows direct regulation of the poses of the trunk and swing foot to avoid overstretched leg joints, enforce a relatively steady trunk posture, and maintain a sufficient clearance between the swing foot and the walking surface.

The desired function $\bm \phi_d( \theta ) $ is parameterized using B\'ezier curves~\cite{westervelt2007feedback}:
	\begin{equation}
	\bm \phi_d( \theta ) 
	:= \sum_{k=0}^{M}\ba_k \frac{M!}{k!(M-k)!}s(\theta)^k(1-s(\theta))^{M-k},
	\label{c5-Bez}
	\end{equation}
where
$M \in \mathbb{Z}^+$ is the order of the B\'ezier curves,
$s(\theta):=\frac{\theta-\theta^+}{\theta^- - \theta^+}$,
$\ba_k \in \mathbb{R}^{10}$ is the unknown vector to be optimized,
and $\theta^+$ and $\theta^-$ are the planned values of $\theta$ at the beginning and the end of a step, respectively.

\noindent \textbf{Optimization formulation.}
The optimization variables are chosen as parameters $\bm \alpha$ in Eq.~\eqref{alpha} and $\ba_k$ in Eq.~\eqref{c5-Bez}.
The constraints are set as:
\begin{enumerate}
	\item[(B1)] 
	The proposed impact invariance conditions (A1)-(A3) in Propositions~\ref{theorem 1} and \ref{theorem1-b}.
	\item [(B2)] Feasibility constraints (e.g., joint-position limits, joint-torque limits,  and ground-contact constraints).
	\item [(B3)] Gait parameters (e.g., step length and duration).
\end{enumerate}
This list of constraints is not intended to be exhaustive as this study focuses on impact invariance construction and controller design instead of trajectory generation. 
MATLAB command $fmincon$ is used to solve the optimization.

\noindent \textbf{Desired trajectories.}
In the simulations and experiments,
the center line $\Gamma_d$ of the desired path is the $X_w$-axis of the world reference frame,
and two desired position trajectories $s_d(t)$ along $\Gamma_d$ are considered, one with a constant velocity and the other with a varying velocity:
\begin{enumerate}
	\item[{a)}] $s_d(t)=4.4t-3$~cm.
	\item[{b)}] $s_d(t)=3.1t-1.5+1.5\sin(0.3t)-\sin(0.8t)$~cm.
\end{enumerate}
The planned virtual constraints are illustrated in Fig.~\ref{fig4-pattern}.

\subsection{Controller Implementation Procedure}
\label{sec-sim-implementation}

This subsection explains the experimental procedure that we adopt to implement the proposed controller on the physical OP3 robot using the ROS package ($op3\_manager$) developed by OP3's manufacturer.

Since the ROS package does not support direct access to the output torques of joint motors,
the proposed control law in Eq.~\eqref{control-law}, which is a torque command, cannot be directly implemented on OP3 and needs to be adapted for its implementation on the robot. 

\begin{figure}[t]
	\centering
	\includegraphics[width=0.75\linewidth]{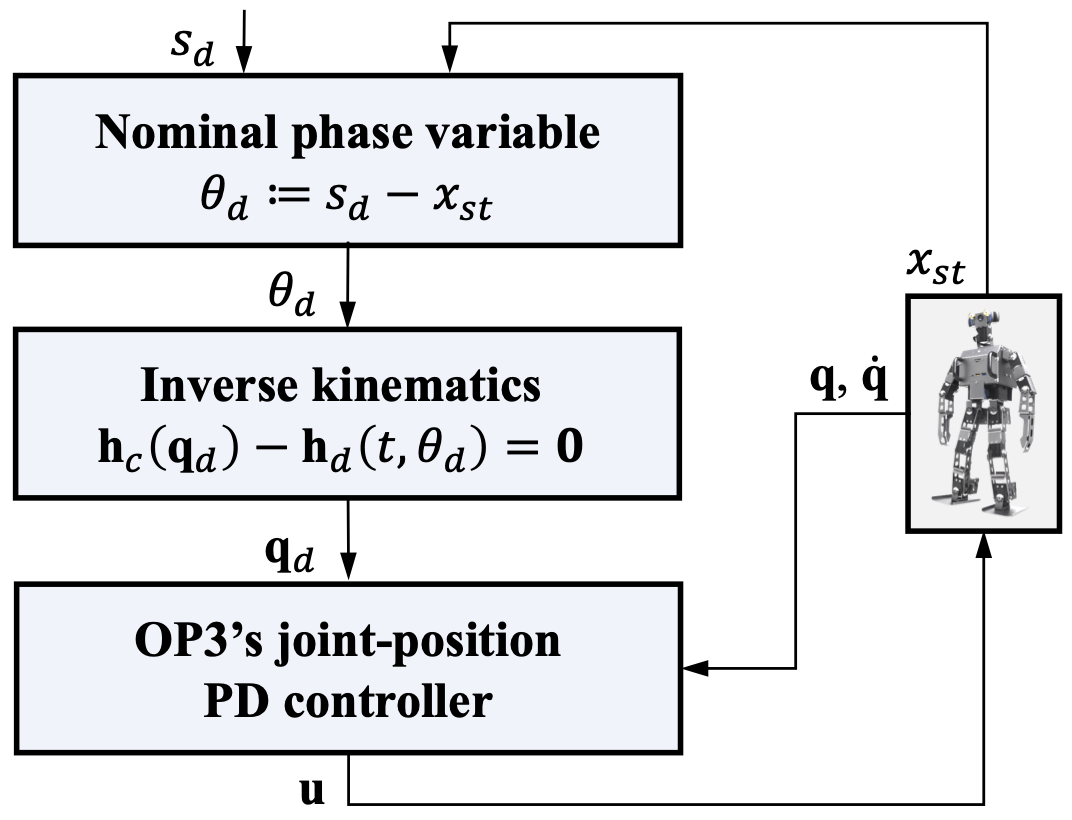}	
	\caption{{Flow chart of the controller implementation procedure for hardware experiments.}
	}
	\label{flow}
\end{figure}

Considering that OP3's ROS package allows users to send desired joint-position trajectories to individual joints and specify the PD gains of OP3's default joint controller, we adopt the following controller implementation procedure~\cite{ames2012dynamically}:
a) to generate the desired position trajectories of individual joints, $\mathbf{q}_d(t)$, and b) to send the desired trajectories to the default joint-position controller.
The main steps of this procedure are shown in Fig.~\ref{flow}.

Although the adapted controller directly tracks the individual joint trajectories $\bq_d$ instead of the original Cartesian-space trajectories $\bh_d$, 
the controller implementation procedure still allows satisfactory tracking of $\bh_d$.
This is because $\mathbf{q}_d$ preserve the feasibility and desired features of $\bh_d$ as specified in (B1)-(B3). 

\subsection{Simulation and Experimental Setup}

\noindent \textbf{MATLAB.}
To validate the theoretical controller design, we utilize MATLAB to implement the control law based on the full-order model of OP3 {(Eq.~\eqref{c5-eq1})}.
The control gains are set as 
$\mathbf{K}_P=225 \cdot  \mathbf{I}$ and $\mathbf{K}_D=30 \cdot   \mathbf{I}$ to ensure the matrix $\bA$ is Hurwitz.
MATLAB simulation results are shown in {Figs.~\ref{fig4} and \ref{fig2-varying}}.

\begin{figure}[t]
	\centering
	\includegraphics[width=1\linewidth]{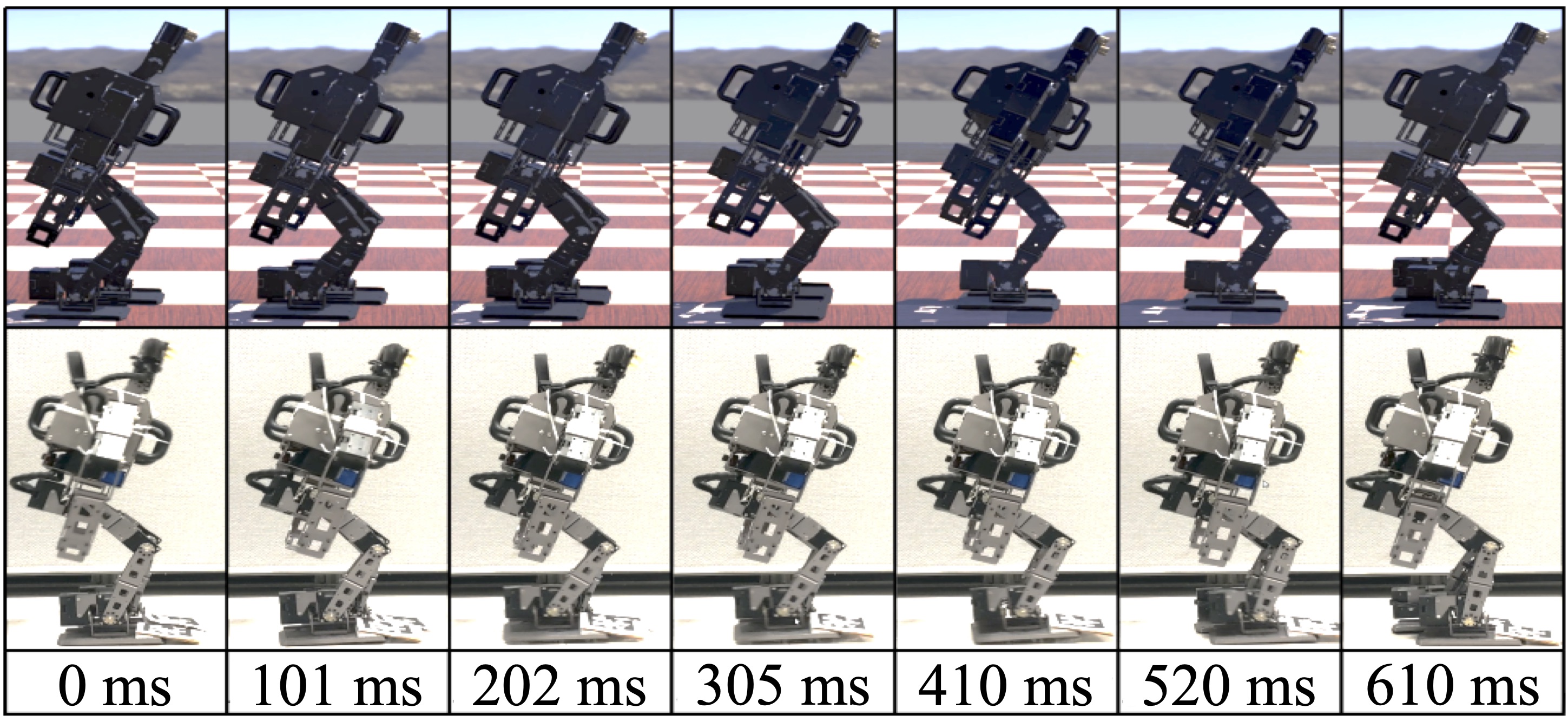}	
	\caption{Time-lapse figures of OP3 walking in Webots simulation (top) and hardware experiment (bottom).
	}
	\label{fig7}
\end{figure}

\noindent \textbf{Webots.}
To gain preliminary insights into the effectiveness of the proposed controller implementation procedure as explained in Section~\ref{sec-sim-implementation},
we use Webots to simulate a 3-D realistic biped model that closely emulates OP3's graphical, physical, and dynamical properties (including its limited actuator accessibility).
The control gains that the emulated robot system allows users to tune are the effective PD gains, whose physical meaning is different from $\bK_P$ and $\bK_D$ in Eq.~\eqref{c5-v}. 
These effective gains are tuned to be ``10'' and ``0'' such that the resulting tracking performance is comparable with the MATLAB results.
Figure~\ref{fig7} shows the time-lapse figures of robot walking obtained in Webots simulations and hardware experiments.
The similarity in the walking gait indicates the validity of using Webots simulations to provide preliminary insights into experiments. 
Webots simulation results of the adapted controller are displayed in Fig.~\ref{fig2}.

\begin{figure}[t]
	\centering 
	\includegraphics[width=0.7\linewidth]{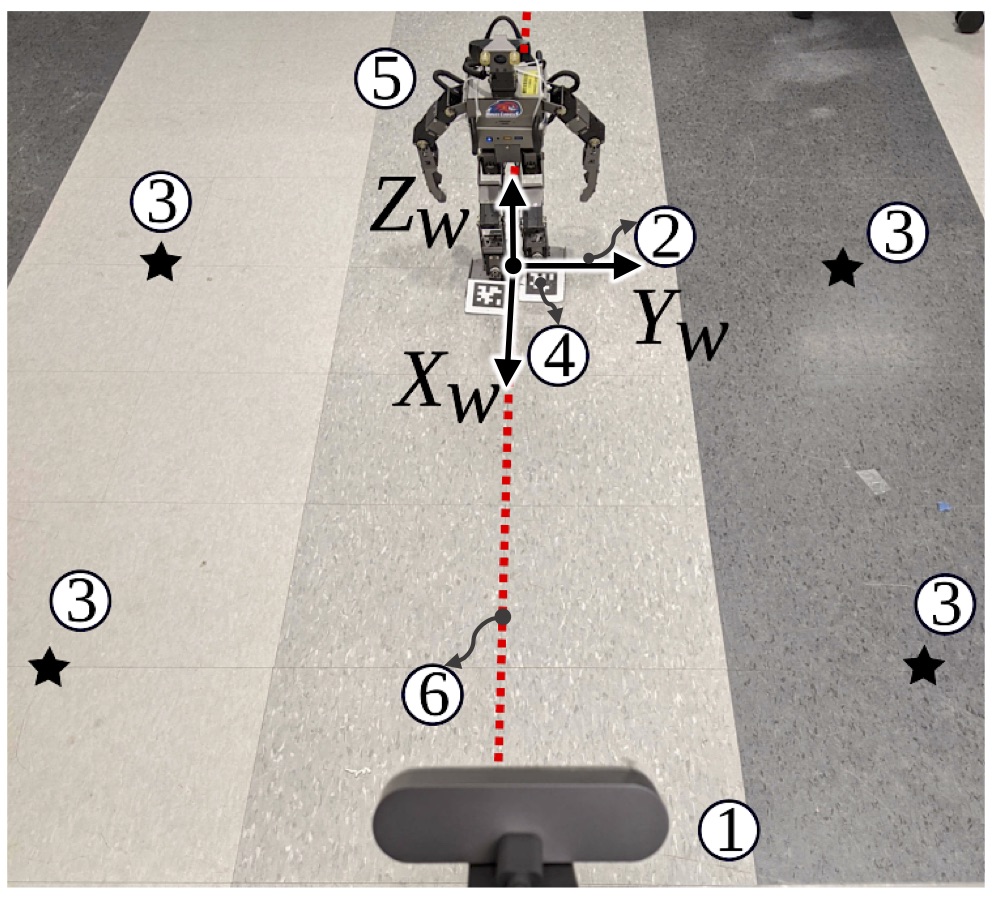}
	\caption{Experimental setup.
		\ding{172}: Logitech 4K PRO WEBCAM.
		\ding{173}: world coordinate frame.
		\ding{174}: reference points for perspective transformation.
		\ding{175}: AprilTag attached to OP3's feet, which is used to determine the robot's global pose.
		\ding{176}: OP3 robot.
		\ding{177}: the center line $\Gamma_d$ of the desired global path.}
	\label{experiment_setup}
\end{figure}

\begin{figure}[t]
	\centering 
	\includegraphics[width=1\linewidth]{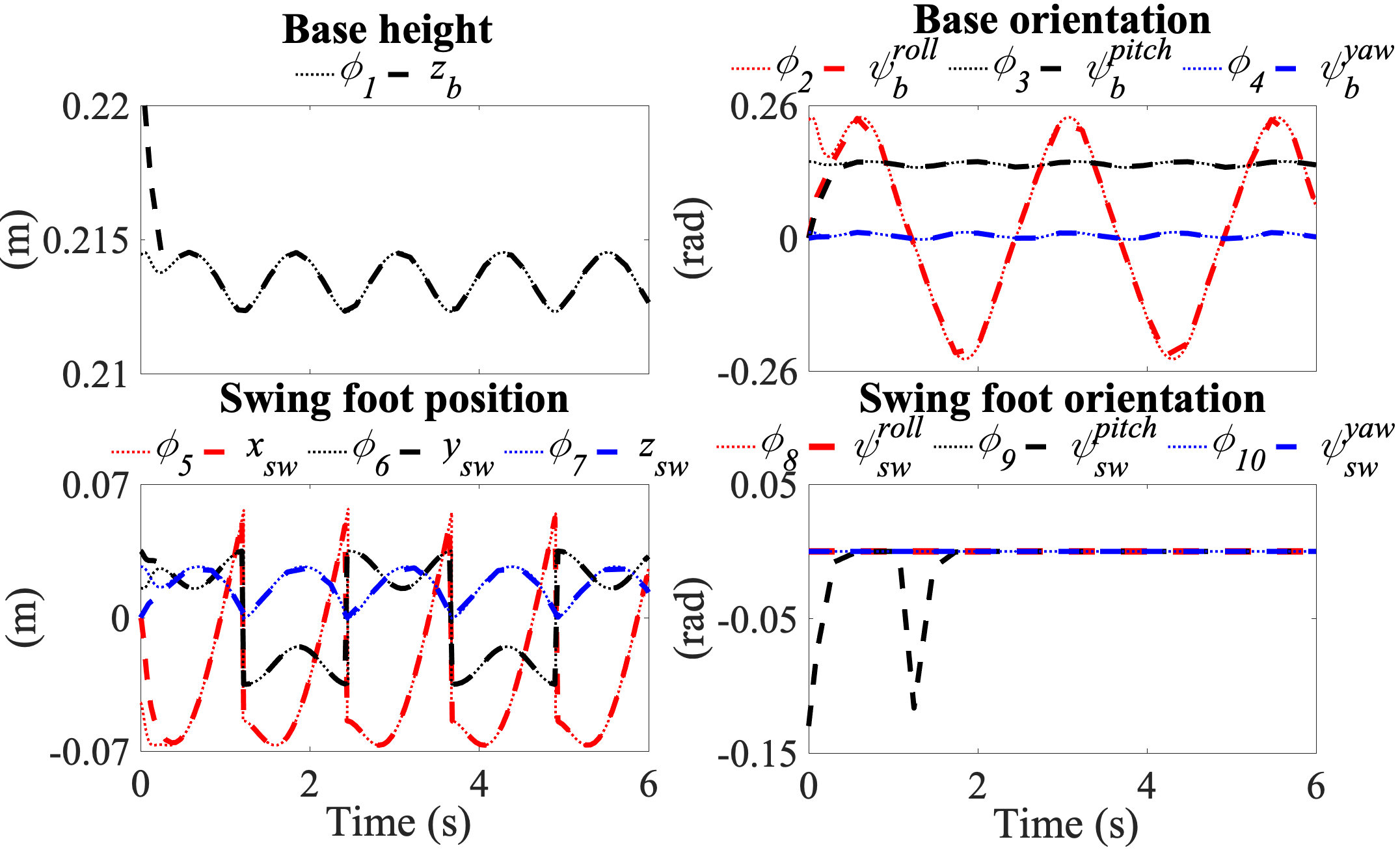}
	\caption{{Asymptotic virtual constraint tracking in MATLAB.
	The functions $\phi_i$ ($i \in \{ 1,2,...10\}$) are elements of the desired function $\bm \phi_d$}.
			} 
	\label{fig4}
\end{figure}

\noindent \textbf{Experiments.}
The experimental setup is shown in Fig.~\ref{experiment_setup}.
With this setup, the robot's joint angles can be directly measured by joint encoders, and its global pose (i.e., position and orientation) can be determined by: a) using the 4K PRO WEBCAM and AprilTag~\cite{olson2011apriltag} to obtain the stance-foot pose in the world reference frame and b) using the obtained stance-foot pose to solve for the robot's global pose via forward kinematics.
By providing relatively accurate measurement, the use of the overhead camera and AprilTag allows us to focus on controller validation.
The experiment is guided by the controller adaptation procedure from Section~\ref{sec-sim-implementation}.
The initial tracking error of the desired position trajectory $s_d$ is $3$~cm, which is approximately $1/3$ of a nominal step length.
The initial path tracking error is $5$ cm.
Similar to the gain tuning in Webots, the effective PD gains are respectively tuned to be ``800'' and ``0'' to ensure a relatively fast error convergence without violating the actuator's torque limit.
Experiment results of OP3 walking on a concrete and a relatively slippery ceramic floor are shown in Fig.~\ref{fig5}.
Videos of the experiments can be accessed at https://youtu.be/VJbLMkOG\_xo.

\begin{figure*}[tp]
	\centering
	\includegraphics[width=1\linewidth]{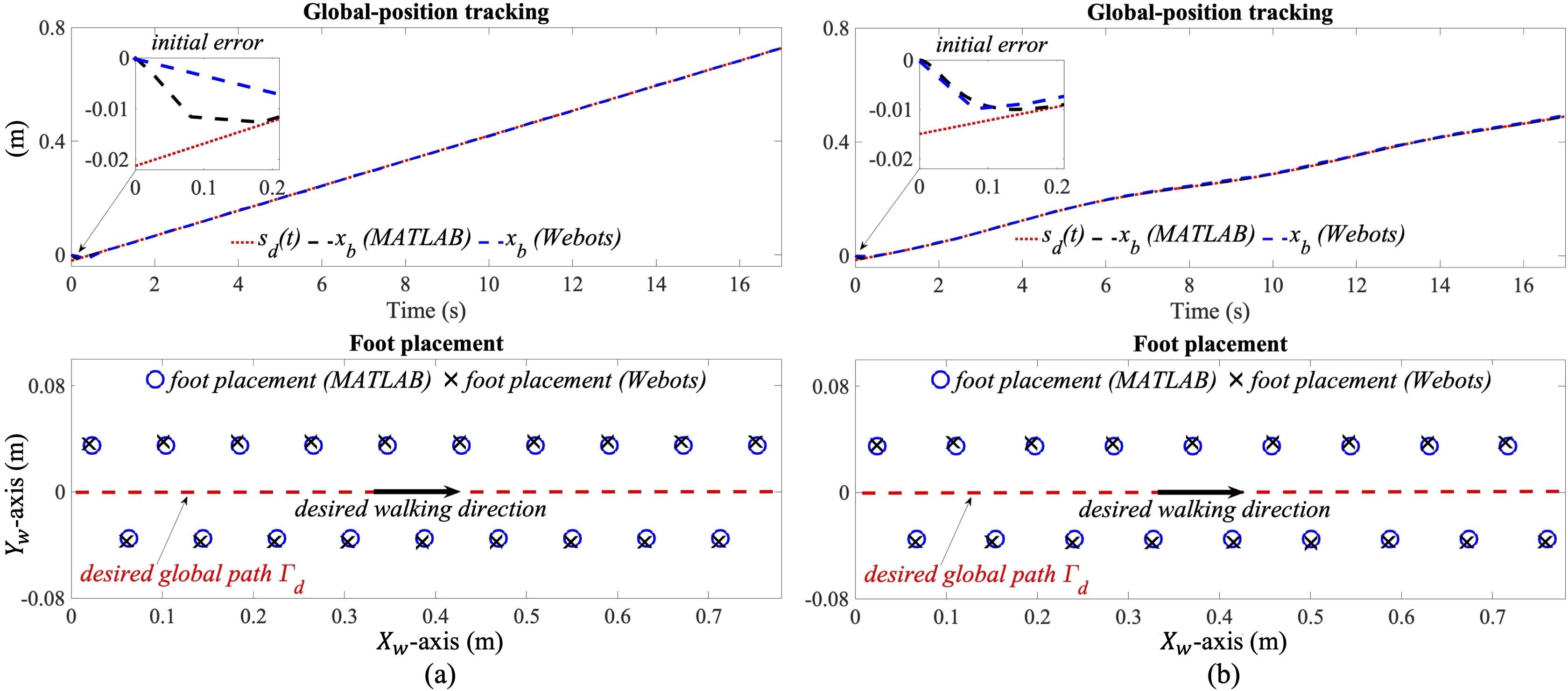}
	\caption{
		{Global-position tracking results in MATLAB and Webots simulations with:
		(a) $s_d(t)=0.044t-0.03$~m and
		(b) $x_d(t)=0.031t-0.015+0.015\sin(0.3t)-0.01\sin(0.8t)$~m.
		}
	}
	\label{fig2}
	\label{fig2-varying}
\end{figure*}

\subsection{Discussions on Validation Results}

\begin{itemize}
\item [\textbf{Tracking accuracy in simulations.}]
The virtual-constraint tracking result in Fig.~\ref{fig4} shows that 
the proposed control law is capable of accurately enforcing the virtual constraints during 3-D fully actuated walking.
The global-position tracking results in Fig.~\ref{fig2} validate that the proposed control law drives the robot to 
asymptotically converge to the desired global-position trajectory $s_d$ while moving along the center line $\Gamma_d$ of the global path.
In particular, the accurate tracking results obtained in Webots indicate the effectiveness of the proposed controller implementation procedure in guaranteeing reliable trajectory tracking in the presence of hardware limitations.
\end{itemize}

\begin{figure*}[t]
	\centering
	\includegraphics[width=1\linewidth]{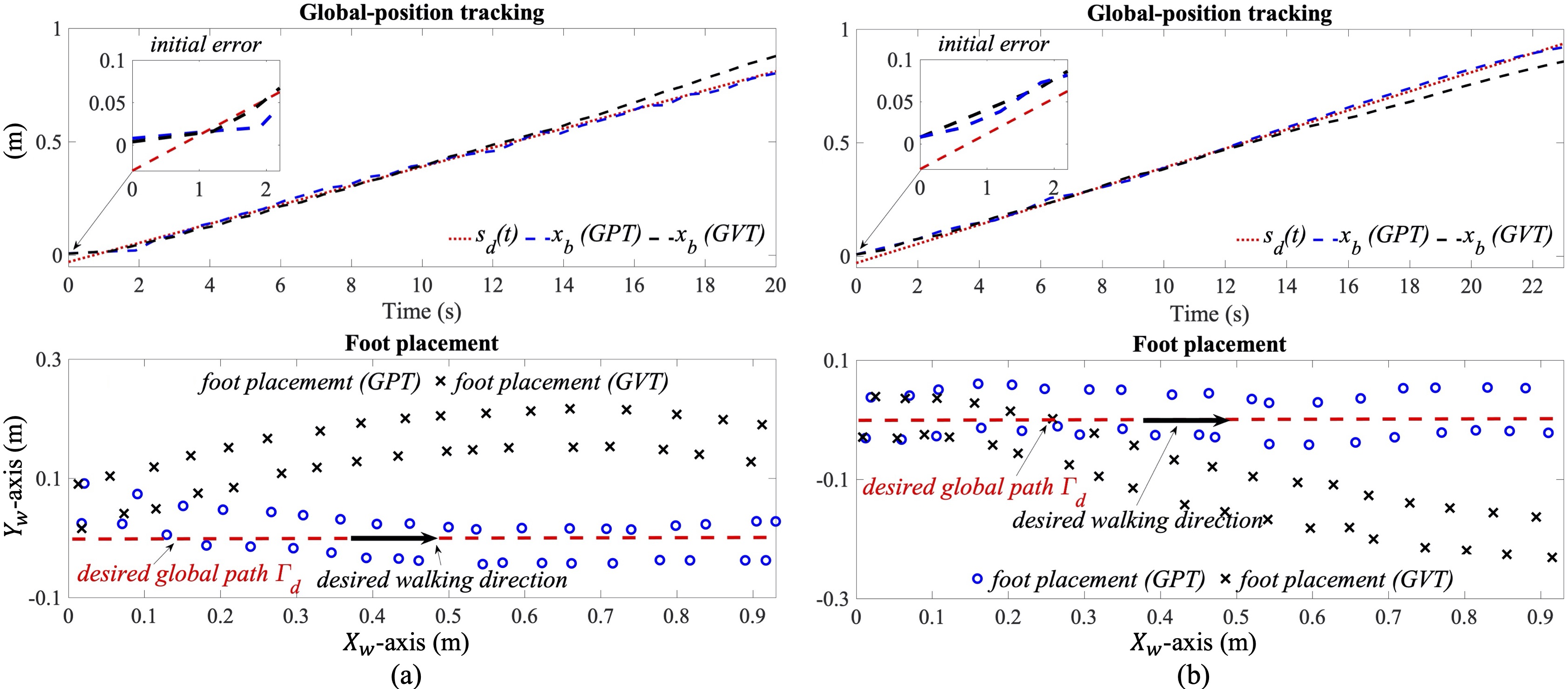} 
	\caption{
		{Experimental results of global-position tracking with $s_d(t)=0.044t-0.03$~m on: (a) a concrete floor and (b) a ceramic tile floor.
		}
		} 
	\label{fig5}
	\label{fig6}
\end{figure*}

\begin{itemize}
\item [\textbf{Tracking accuracy in experiments.}]
As illustrated in Fig.~\ref{fig5} (a) (top), under the proposed global-position tracking (GPT) controller, the robot's actual global position $x_{b}$ (labeled as {``$x_b$ (GPT)''}) converges to a relatively small neighborhood about its desired trajectory $s_d$ within $3$ seconds when the robot walks on a concrete floor.
Also, Fig.~\ref{fig5} (a) (bottom) illustrates that despite an initial path tracking error of $5$ cm, the robot remains close to the center line $\Gamma_d$ of the desired global path, as indicated by the footstep trajectories labeled as {``\it foot placement (GPT)''}.
Due to uncertainties such as hardware limitations, modeling errors, and floor surface irregularity, achieving an exactly zero steady-state tracking error on a physical robot may not be feasible. 
Thanks to the inherent robustness of feedback control, the proposed control approach achieves a small steady-state tracking error, although uncertainties are not explicitly addressed in the proposed theoretical controller design.
\end{itemize}

\begin{itemize}
    \item [\textbf{Robustness.}]
To further test the limit of the inherent robustness of the proposed control approach, experiments of OP3 walking on a ceramic tile floor were conducted (Fig.~\ref{fig6} (b)).
As the surface of the ceramic tiles is relatively more slippery than the concrete floor, the robot's stance foot slips more frequently on the tile floor, causing a stronger violation of the modeling assumption of static stance foot. 
{Yet, a relatively small global-position tracking error is still realized when the initial foot placement error is small, as shown in Fig.~\ref{fig5}(b).}
\end{itemize}

\begin{itemize}
\item [\textbf{Comparison with global-velocity tracking}]  \textbf{control.}
Results of a global-velocity tracking {(GVT)} controller, which is analogous to the orbitally stabilizing controller for 3-D fully actuated walking~\cite{ames2012dynamically}, are also displayed in {Figs.~\ref{fig5}}.
Although the GPV controller achieves accurate tracking of the desired global velocity $\dot{s}_d$,
its global trajectory tracking performance is not satisfactorily guaranteed, as indicated by the relatively large deviations of the global position (labeled as ``{$x_b$ (GVT)}'') and the footstep trajectories (labeled as ``{\it foot placement (GVT)}'').
\end{itemize}

\section{Discussions}
\label{sec-discussions}

This study has extended the previous method of impact invariance construction from orbital stabilization to the stabilization~\cite{westervelt2003hybrid} of time-varying global-position trajectory during 3-D walking.
The proposed method produces impact invariance conditions that can be imposed in the trajectory generation of virtual constraints for ensuring their agreement with impact dynamics. 
Moreover, although the impact maps of the virtual constraints and global trajectory are generally nonlinearly coupled through the robot's kinematic chains,
these conditions can automatically ensure any arbitrary smooth desired global-position trajectory respects the impact dynamics.
Indeed, as shown in Fig.~\ref{fig2}, the proposed controller achieves asymptotic tracking of two different global-position trajectories under the same virtual constraints (Fig.~\ref{fig4-pattern}), indicating that the virtual constraints ensure the impact agreement for different desired global-position trajectories. 
Thus, the proposed impact invariance conditions can allow the decoupling between the lower-level trajectory generation of virtual constraints and the higher-level planning of global-position trajectory.
The decoupling could permit offline planning of virtual constraints, thus reducing the computational burden of motion planning.


This study has also introduced the Lyapunov-based stability conditions
for the hybrid closed-loop error system associated with 3-D walking. 
Controller designs satisfying these conditions can accurately track the time-varying desired global-position trajectory, as demonstrated in Figs.~\ref{fig2} and \ref{fig5}.
The proposed control approach can also indirectly drive the lateral foot placement $y_{st}$ to the desired location $y_{std}$,
which is predicted by the asymptotic convergence of the Lyapunov function $V_a$ that explicitly contains the lateral foot-placement error.
Note that our previous controller for 2-D walking cannot address the convergence of $y_{st} - y_{std}$
as it does not consider the robot's lateral movement.
The capability of accurate foot placement could potentially be exploited to handle locomotion on discrete terrains (e.g., stepping stones~\cite{nguyen2020dynamic}).

Uncertainties such as modeling errors and terrain irregularities are prevalent during real-world robot operations~\cite{gaathon2020minimalistic}.
The proposed control approach achieves a small final tracking error when the uncertainties (e.g., walking surface irregularities) are relatively small, as demonstrated by the experiment results in Fig.~\ref{fig5}.
However, if the uncertainties are significant, the controller may not guarantee a reliable tracking performance because it does not explicitly deal with uncertainties.
One potential approach to improve robustness is to integrate the proposed control law with adaptive and robust control~\cite{liao2017high,yuan2019fast} for enabling online model estimation and better disturbance rejection. 

Real-world applications of legged robots commonly require walking in varying directions.
To apply and extend the proposed approach from straight-line to curved-path locomotion,
the impact invariance construction method can be directly incorporated in virtual constraint planning to ensure impact agreement.
Also, to allow efficient planning, we can construct a library~\cite{da2016from} of virtual constraints offline that correspond to a common range of direction-varying gait parameters, and interpolate the virtual constraints online to fit the varying walking directions during curved-path navigation.

\section{Conclusions}
\label{sec-conclusions}
This paper has introduced a control approach that explicitly addresses the hybrid robot dynamics for achieving asymptotic global-position tracking during fully actuated 3-D bipedal walking.
With the output function designed as the tracking error of the desired global-position trajectory and virtual constraints,
a continuous input-output linearizing control law was synthesized to asymptotically drive the output function to zero within continuous phases.
The construction of impact invariance conditions was introduced to inform the generation of virtual constraints such that the robot's desired motions defined by the virtual constraints and the desired global-position trajectory all respect the discrete landing impact dynamics.
Sufficient conditions were derived based on Lyapunov theory under which the proposed continuous control law provably guarantees the asymptotic tracking performance of the hybrid closed-loop system.
Simulation and experimental results demonstrated the effectiveness of the proposed control approach in realizing satisfactory global-position tracking during 3-D walking.

\bibliographystyle{asmems4}        
\bibliography{asme}  

\section{Appendix: Proofs of All Propositions and Theorems}

\label{proof}

\subsection{Proof of Proposition~\ref{prop2}}

With the pre-impact joint position $\bq^*$,
the post-impact joint position and phase variable are
$
\bq(\tau_k^+) 
= \bm \Delta_q (\bq^*) = \bq_0
$
and
$
\theta(\tau_k^+) 
= \bar{x}_b(\bq(\tau_k^+))
=\bar{x}_b(\bq_0)
=\theta_0
$,
respectively.

Under the foot-placement condition $y_{st}=y_{std}$ and condition (A1), the post-impact value of the output function $\bar{y}_b -(y_d- y_{st})$ is
$
\bar{y}_b ( \bq_0    )  -   y_d ( \theta_0 ) + y_{st}(\tau_k^+)
=
\bar{y}_b ( \bq_0    )  -   y_d ( \theta_0 ) + y_{std}  =0.
$
Similarly, given condition (A1), the post-impact value of $\bm \phi_c- \bm \phi_d$ is
$
\bm \phi_c(\bq_0) - \bm \phi_d ( \theta_0 ) = \bzero.
$
Thus, the impact invariance of the output functions $\bar{y}_b -(y_d- y_{st})$ and $\bm \phi_{c}-\bm \phi_d$ is ensured.

Since $\dot{\by}(\tau_k^-)=\bzero$, we have
\begin{equation}
\begin{aligned}
\dot{\by}(\tau_k^-)
&= 
\dot{\bh}_c(\bq^*,\dot{\bq}(\tau_k^-))
-
\dot{\bh}_d(t,\theta^*,\dot{\theta}(\tau_k^-))
\\
&=
\bJ_{h_c} ( \bq^*)  \dot{\bq}(\tau_k^-)
-
\begin{bmatrix}
\dot{s}_d(\tau_k^-) \\
\frac{ \partial y_d  }{ \partial \theta} ( \theta^* ) \dtheta(\tau_k^-)   \\
\frac{ \partial \bm \phi_d   }{ \partial \theta}( \theta^* )  \dtheta(\tau_k^-)  \\
\end{bmatrix} 
=\bzero   
\end{aligned}
\end{equation} 
and 
$
\dot{\theta}(\tau_k^-) 
=
\frac{\partial \bar{x}_b  }{\partial \bq} (\bq^*) \dot{\bq}(\tau_k^-) 
=
\dot{s}_d(\tau_k^-) - \dot{x}_{st}(\tau_k^-)
= \dot{s}_d(\tau_k^-),
$
where $\bJ_{h_c}(\bq^*):= \frac{ \partial {\bh}_c }{ \partial \bq  } (\bq^*)$.
Thus, the pre-impact joint velocity is:
\begin{equation}
\dot{\bq}(\tau_k^-)
=
\bJ_{h_c}^{-1} ( \bq^*)  
\begin{bmatrix}
1 \\
\frac{ \partial y_d  }{ \partial  \theta} ( \theta^* ) \\
\frac{ \partial \bm \phi_d   }{ \partial  \theta}( \theta^* ) \\
\end{bmatrix} \dtheta(\tau_k^-) .
\label{dq_plus}
\end{equation}
Note that
$
\dot{\bq}  (\tau_k^+) = \bm \Delta_{\dq} (\bq^*)  \dot{\bq} (\tau_k^-)
$
and
$
\dtheta (\tau_k^+)  = \frac{ \partial \bar{x}_b }{ \partial \bq  }  (    \bq_0   ) \dot{\bq} (\tau_k^+)
$.
Then, by condition (A2), 
the post-impact value of
$
\begin{bmatrix}
\dot{\bar{y}}_b - (\dot{y}_d - \dot{y}_{st}) \\
\dot{\bm \phi}_c - \dot{\bm \phi}_d
\end{bmatrix}
$ 
becomes
$
\begin{bmatrix}
\frac{ \partial \bar{y}_b}{\partial \bq} (\bq_0) \\
\frac{\partial \bm \phi_{c}}{\partial \bq}	(\bq_0)
\end{bmatrix}
\dot{\bq} (\tau_k^+) 
		-
\begin{bmatrix}
\frac{ \partial y_d  }{ \partial \theta} ( \theta_0  )  
\\
\frac{ \partial \bm \phi _d  }{ \partial \theta} ( \theta_0) 
\end{bmatrix}
\dot{\theta} (\tau_k^+) 
=
\bzero
$;
that is, the impact invariance of $
\begin{bmatrix}
\dot{\bar{y}}_b - (\dot{y}_d - \dot{y}_{st}) \\
\dot{\bm \phi}_c - \dot{\bm \phi}_d
\end{bmatrix}
$ is met.

\subsection{Proof of Proposition~\ref{theorem1-b}}

Because $x_b$ and $s_{d}(t)$ are both continuous in $t$,
we obtain
\begin{equation*}
		\bar{x}_b ( \bq_0    )  -   s_d ( \tau_k^+ ) + x_{st}
		=
		{x}_b ( \tau_k^+   )  -   s_d ( \tau_k^+ )
		=
		{x}_b ( \tau_k^-   )  -   s_d ( \tau_k^- ).
		\label{c5-eq39}
\end{equation*}
Thus, if ${x}_b ( \tau_k^-   )  -   s_d ( \tau_k^- )=0$, then $\bar{x}_b ( \bq_0    )  -   s_d ( \tau_k^+ ) + x_{st}=0$;
that is, the impact invariance of $x_b - s_d$ automatically holds.

Because the stance foot remains static just before and after the impact,
$\dot{x}_{st}(\tau_k^+)=\dot{x}_{st}(\tau_k^-)=0$ holds.
Also, as the desired global velocity $\dot{s}_d$ is continuous in $t$, we have
$\dot{s}_{d} (\tau_k^+) = \dot{s}_{d} (\tau_k^-)$.

From the proof of Proposition~\ref{theorem 1}, we have
	\begin{equation}
	\begin{aligned}
	    \dtheta (\tau_k^+)  
		&=
		\frac{ \partial \bar{x}_b }{ \partial \bq  }  (    \bq_0   )
		\bm \Delta_{\dq} (\bq^*)  \bJ_h^{-1} ( \bq^* )
			\begin{bmatrix}
				1
				\\
				\frac{ \partial y_d  }{ \partial \theta} ( \theta^* ) 
				\\
				\frac{ \partial \bm \phi_d  }{ \partial \theta} ( \theta^* ) 
			\end{bmatrix}  
			\dtheta (\tau_k^-),
	\end{aligned}
		\label{c5-eq45}
	\end{equation}
which yields $	 \dtheta (\tau_k^+)	= \dtheta (\tau_k^-)$ under condition (A3).
Then,
the post-impact value of the first derivative of the output function $x_b- s_d$ becomes
$\dot{\bar{x}}_b ( \bq_0, \dot{\bq} (\tau_k^+) )  - (  \dot{s}_d ( \tau_k^+ ) - \dot{x}_{st}^+  ) =
\dtheta (\tau_k^+)  -   \dot{s}_d ( \tau_k^+ ) + \dot{x}_{st}^+
= \dtheta (\tau_k^-)  -   \dot{s}_d ( \tau_k^- ) + \dot{x}_{st}^-
$,
which is zero if $ \dtheta (\tau_k^-)  -   \dot{s}_d ( \tau_k^- ) + \dot{x}_{st}^-=0$.
Thus, the impact invariance of the global-position tracking error state holds.

\subsection{Proof of Proposition~\ref{prop-t-bound}}

Because the output function state $\by$ and $\dot{\by}$ and the swing-foot height $z_{sw}$ defining the switching surface $S$ are both continuously differentiable in their respective arguments, the function defining the switching surface $S_x$ is continuously differentiable in its argument~\cite{gu2017exponential}.
Also, note that the continuous-phase vector field (i.e., $\bA \bx$) of the error state $\bx$ is continuously differentiable in $\bx$.

Then, by Lemma 2.1 and Corollary 2.4 in~\cite{gu2017exponential}, the impact timing $T_k$ is an implicit function of the state $\bx$, and is Lipschitz continuous with respect to $\bx$.
Thus, there exists a positive number $r_1$ and a Lipschitz constant $L_{T_x}$ such that
$
| T_k - \tau_k | \leq L_{T_x} \|  \tilde{\bx}(\tau_k;  T_{k-1}^+,\bx|^+_{k-1} )   \| 
$
for any $\bx|_{0}^+ \in B_{r_1} (\bzero)$ and any $k \in \mathbb{Z}^+$.

\subsection{Proof of Proposition~\ref{prop-y-bound}}

Let $\phi_{sw,y}(\theta)$ denote the desired trajectory of the control variable ${y}_{sw}$.
Because the stance-foot position during the $(k+1)^{th}$ step is the swing-foot position at the end of the $k^{th}$ step,
one has $y_{st}|_{k}^+ = {y}_{sw}|_{k}^-$
and $\phi_{sw,y}(\theta^*) = y_{std}$.

Let $\tilde{\theta}(t;T_{k-1}^+,\bx|^+_{k-1})$ be the phase variable associated with the fictitious state $\tilde{\bx}(t;T_{k-1}^+,\bx|^+_{k-1}) $ that overlaps with the actual state $\bx$ on $t \in [T_{k-1}^+,T_{k}^-]$.
Then, by the triangular inequality, we can approximate the upper bound of the absolute lateral foot-placement error as
	\begin{equation}
		\begin{aligned}
			&~ |  y_{st}|_{k}^+ - y_{std} |
			= | {y}_{sw}|_{k}^-  - \phi_{sw,y}(\theta^*) |  \\
			\leq
			&~ | {y}_{sw}|_{k}^-  -  \phi_{sw,y} (\theta(T_{k}^-))  | \\ 
			& + | \phi_{sw,y} (\theta(T_{k}^-)) - \phi_{sw,y}(  \tilde{\theta}(\tau_k;T_{k-1}^+,\bx|^+_{k-1})  ) | \\ 
			& + | \phi_{sw,y}(  \tilde{\theta}(\tau_k;T_{k-1}^+,\bx|^+_{k-1})  ) - \phi_{sw,y} (\theta^*)|.
		\end{aligned}
		\label{c5-eq32}
	\end{equation}
The upper bounds of the three terms on the right-hand side of this inequality are derived next.

As ${y}_{sw}-\phi_{sw,y}$ is an element of the full error state $\bx$, its norm satisfies
	\begin{equation}
		| {y}_{sw}|_{k}^-  -  \phi_{sw,y} (\theta(T_{k}^-))  | \leq \| \bx |_k^- \|.
	\end{equation}
	
Because$\theta(T_{k}^-) =  \tilde{\theta}(T_k^-;T_{k-1}^+,\bx|^+_{k-1})  $ and because $\phi_{sw,y}(\theta)$ and 
$\tilde{\theta}(t;T_{k-1}^+,\bx|^+_{k-1})  $
are continuously differentiable in $\theta$ and $t$, respectively,
there exists a positive number $r_2$ and Lipschitz constants $L_{\phi_{sw,y}} $ and $L_{\theta_t} $ such that 
\begin{equation}
\begin{aligned}
			&~\|  \phi_{sw,y} (\theta(T_{k}^-)) - \phi_{sw,y}(  \tilde{\theta}(\tau_k;T_{k-1}^+,\bx|^+_{k-1})  )  \| 
			\\
			\leq &~L_{\phi_{sw,y}} 
			\| \theta(T_{k}^-) -  \tilde{\theta}(\tau_k;T_{k-1}^+,\bx|^+_{k-1})  \|  
			\\
			=&~ L_{\phi_{sw,y}} 
			\|   \tilde{\theta}(T_k^-;T_{k-1}^+,\bx|^+_{k-1})    -  \tilde{\theta}(\tau_k;T_{k-1}^+,\bx|^+_{k-1})  \|
			\\
			\leq &~L_{\phi_{sw,y}}  L_{\theta_t}  | T_k - \tau_k | 
\end{aligned}
\label{c5-eq33}
\end{equation}
for any $\bx|_{0}^+ \in B_{r_2} (\bzero) := \{  \bx \in \mathbb{R}^{2n} :  \| \bx \| \leq r_2    \}$.
	
With $\theta^*  = \theta(\bq^*) =  \tilde{\theta}(\tau_k^-;T_{k-1}^+,\bzero) = s_{d}(\tau_k^-) - x_{st}(\tau_k^-),$
we have
	\begin{equation}
		\begin{aligned}
			&~| \tilde{\theta}(\tau_k;T_{k-1}^+,\bx|^+_{k-1}) - \theta^* | 
			\\
			=&~
			\|    \tilde{\theta}(\tau_k;T_{k-1}^+,\bx|^+_{k-1})   -    (s_{d}(\tau_k^-) - x_{st}(\tau_k^-)   )   \| 
			\\
			\leq&~
			\| \tilde{\bx}(\tau_k;  T_{k-1}^+,\bx|^+_{k-1} )  \|.
			\label{c5-eq36}
		\end{aligned}
	\end{equation}
Then,
	\begin{equation}
	    \begin{aligned}
	        &| \phi_{sw,y}(  \tilde{\theta}(\tau_k;T_{k-1}^+,\bx|^+_{k-1})  ) - \phi_{sw,y} (\theta^*)| 
	        \\
	\leq & L_{\phi_{sw,y}} \|  \tilde{\theta}(\tau_k;T_{k-1}^+,\bx|^+_{k-1}) -  \theta^* \|
	\leq 
	L_{\phi_{sw,y}} \| \tilde{\bx}(\tau_k;  T_{k-1}^+,\bx|^+_{k-1} )  \|.
	    \end{aligned}
	    \label{c5-eq333}
	\end{equation}
Let $\beta_{st} :=  L_{\phi_{sw,y}}( L_{\theta_t} L_{T_x}  + 1 )$ and $d_1 := \min (r_1,r_2)$.
From Proposition~\ref{prop-t-bound} and Eqs.~\eqref{c5-eq32}-\eqref{c5-eq333}, we obtain
$
		| y_{st}|_{k}^+ - y_{std} | \leq  \| \bx|^-_{k} \| + \beta_{st}  \|  \tilde{\bx}(\tau_k;T_{k-1}^+, \bx|^+_{k-1}) \| 
$
for any $\bx|_{0}^+ \in B_{d_1} (\bzero)$ and $k \in \mathbb{Z}^+$.

\subsection{Proof of Theorem~\ref{theorem2}}

\label{proof-thm2}

To simplify the stability analysis using the proposed conditional impact invariance, which holds when $y_{st}=y_{std}$, we will explicitly analyze the convergence of $y_{st}$ to $y_{std}$.
An augmented Lyapunov function candidate is then constructed as
$
V_a(\bx,  y_{st} - y_{std}):=V(\bx) + \sigma  (y_{st} - y_{std})^2,
$
where $\sigma$ is a positive number to be specified later.

By the stability theory based on the construction of multiple Lyapunov functions~\cite{branicky1998multiple},
the origin of the hybrid time-varying system in Eq.~\eqref{c5-hybrid} is locally asymptotically stable if there exists a positive number $d_2$ such that
for any $\bx|_0^+ \in B_{d_2} (\bzero)$, $V_a$ is monotonically decreasing within each continuous phase and $\{V_a |^+_{1}, V_a|^+_{2}, V_a|^+_{3} ... \}$ is a strictly decreasing sequence with $V_a |^+_{k} \rightarrow 0$ as $k \rightarrow \infty$.
	
\noindent \textbf{Evolution of $V_a$ during continuous phases.} With the PD gains chosen such that $\bA$ is Hurwitz, Eq.~\eqref{c5-eq17} gives
$
V|^- _{k} \leq e^{- \frac{c_3}{c_2} (T_{k+1} - T_{k})} V|^+_{k-1}
\label{c5-Lyap2}
$
within the $k^{th}$ ($k \in \mathbb{Z}^+$) continuous phase.
Since $y_{st}-y_{std}$ remains constant within the step due to the static stance foot,
$V_a$ monotonically decreases within the $k^{th}$ phase.

\noindent \textbf{Evolution of $V_a$ across nonlinear impact maps.}
Consider the foot-landing event at the end of the $k^{th}$ walking step (i.e., $t=T_k^-$).
The tracking error expansion across the landing event is analyzed as follows.

Because the desired functions $y_d$ and $\bm \phi_d (\theta)$ satisfy the conditions (B1)-(B3), the impact invariance of the error state $\bx$ holds, which leads to $ \bm \Delta (\tau_{k}^-, \bzero, y_{std}) = \bzero$.
Then, the value of $\bx$ just after the landing can be approximated by applying the triangular inequality as:
\begin{equation}
		\begin{aligned}
			\| \bx|^+_{k} \| 
			= &~ \| \bm \Delta(T_{k}^-, \bx|^-_{k}, y_{st}|_{k}^-  )  \| 
			\\
			= &~ \| \bm \Delta(T_{k}^-, \bx|^-_{k}, y_{st}|_{k}^-  ) - \bm \Delta (\tau_{k}^-, \bzero, y_{std}  )     \|
			\\
			\leq
			&~ 
			\| \bm \Delta(T_{k}^-, \bx|^-_{k}, y_{st}|_{k}^-  )  -  \bm \Delta (\tau_{k}^-, \bx|^-_{k}, y_{st}|_{k}^-  )     \| 
			\\
			+& 
			\| \bm \Delta (\tau_{k}^-, \bx|^-_{k}, y_{st}|_{k}^- )  -  \bm \Delta (\tau_{k}^-, \bzero, y_{st}|_{k}^- )     \| 
			\\
			+& 
			\| \bm \Delta (\tau_{k}^-,  \bzero, y_{st}|_{k}^-  )  -  \bm \Delta (\tau_{k}^-, \bzero, y_{std}  )     \| 
			+ \|   \bm \Delta (\tau_{k}^-, \bzero, y_{std}  )     \| .
		\end{aligned}
		\label{c5-eq49}
\end{equation} 
As the reset map $\bm \Delta  (t, \bx, y_{st}  ) $ is continuously differentiable in $t$, $\bx$, and $y_{st}$,
there exists a positive number $r_3$ and Lipschitz constants $L_{\Delta_t}$, $L_{\Delta_x}$, and $ L_{\Delta_{st}} $ such that the following inequalities hold for any $\bx|_{0}^+ \in B_{r_3} (\bzero)$:
\begin{equation}
		\begin{aligned}
			&\| \bm \Delta (T_{k}^-, \bx|^-_{k}, y_{st}|_{k}^-  )  -  \bm \Delta (\tau_{k}^-, \bx|^-_{k}, y_{st}|_{k}^- )     \| \leq L_{\Delta_t} | T_{k} - \tau_{k}  | .
			\\
			& \| \bm \Delta (\tau_{k}^-, \bx|^-_{k},y_{st}|_{k}^- )  -  \bm \Delta (\tau_{k}^-, \bzero, y_{st}|_{k}^- )     \|  \leq L_{\Delta_x} \| \bx|^-_{k} \|.
			\\
			& \| \bm \Delta (\tau_{k}^-,  \bzero, y_{st}|_{k}^- )  -  \bm \Delta (\tau_{k}^-, \bzero, y_{std}  )     \| 
			\leq  L_{\Delta_{st}}  | y_{st}|_{k}^- - y_{std} | .
		\end{aligned}
		\label{c5-Lip}
	\end{equation}
From Eqs.~\eqref{c5-T} and \eqref{c5-Lip}, we obtain
	\begin{equation}
		\begin{aligned}
			&~\| \bm \Delta (T_{k}^-, \bx|^-_{k}, y_{st}|_{k}^- )  -  \bm \Delta (\tau_{k}^-, \bx|^-_{k}, y_{st}|_{k}^-  )     \|
			\\
			\leq& ~L_{\Delta_t} L_{T_x} \|  \tilde{\bx}(\tau_k;  T_{k-1}^+, \bx|^+_{k-1} )   \| 
		\end{aligned}
		\label{c5-eq52}
	\end{equation}
for any $\bx|_{0}^+ \in B_{d_2} (\bzero)$, where $d_2 = \min\{d_1,r_3\}$.
From Eqs.~\eqref{c5-eq49} - \eqref{c5-eq52}, we have for any $\bx|_0^+ \in B_{d_2} (\bzero)$, 
	\begin{equation}
		\begin{aligned}
			&\| \bx|^+_{k} \| 
			= \| \bm \Delta (T_{k}^-, \bx|^-_{k}, y_{st}|_{k}^-  )  \| 
			\\
			\leq ~& 
			L_{\Delta_x}    \| \bx|^-_{k} \|
			+
			L_{\Delta_t}  L_{T_x}  \|  \tilde{\bx}(\tau_k;T_{k-1}^+, \bx|^+_{k-1}) \| 
			+
			L_{\Delta_{st}}  | y_{st}|_{k}^- - y_{std} |  . 
		\end{aligned}
		\label{c5-x_plus}
	\end{equation}
The upper bounds of
$\| \bx|^-_{k} \|$ and
$\|  \tilde{\bx}(\tau_k;T_{k-1}^+, \bx|^+_{k-1}) \|$
with respect to the tracking error norm $\| \bx|^-_{k-1} \|$ can be derived based on Eq.~\eqref{c5-eq17} as:
\begin{equation}
		\begin{aligned}
			&\| \bx|^-_{k} \| \leq \sqrt{\tfrac{c_2}{c_1}} e^{-\frac{c_3}{2 c_2} (T_{k} - T_{k-1}) } \| \bx|^+_{k-1} \|~\mbox{and}~
			\\
			& \|  \tilde{\bx}(\tau_k;T_{k-1}^+, \bx|^+_{k-1}) \|
			\leq 
			\sqrt{\tfrac{c_2}{c_1}} e^{-\frac{c_3}{2 c_2} (\tau_{k} - T_{k-1}) } \| \bx|^+_{k-1} \| .
		\end{aligned}
		\label{c5-x_con}
\end{equation}
Then, from Eqs.~\eqref{c5-x_plus} and \eqref{c5-x_con}, the post-impact error norm can be approximated as:
\begin{equation}
		\begin{aligned}
			\| \bx|^+_{k} \| 
			\leq
			&
			\sqrt{\tfrac{c_2}{c_1}}  
			(L_{\Delta_t}  L_{T_x} + L_{\Delta_x}  e^{-\frac{c_3}{2 c_2} (T_{k} - \tau_{k}) }  )
			e^{-\frac{c_3}{2 c_2} (\tau_{k} - T_{k-1}) } 
			\| \bx|^+_{k-1} \|
			\\
			&+
			L_{\Delta_{st}}  | y_{st}|_{k}^- - y_{std} | .
		\end{aligned}
		\label{c5-x+}
	\end{equation}
	\normalsize
For any $\epsilon>0$ there exist PD gains corresponding to a sufficiently high convergence rate $\frac{c_3}{2 c_2} $ such that
$
		e^{-\frac{c_3}{2 c_2} (T_{k} - \tau_{k}) }  \leq 1+\epsilon.
		\label{c5-epsilon}
$
Then, the approximation of the post-impact error norm can be simplified into
	\begin{equation}
		\| \bx|^+_{k} \| 
		\leq
		\alpha_x
		\| \bx|^+_{k-1} \|
		\\
		+
		\alpha_{st}
		| y_{st}|_{k}^- - y_{std} |,
		\label{x-plus}
	\end{equation}
	where $\alpha_x := \sqrt{\frac{c_2}{c_1}}   (L_{\Delta_t}  L_{T_x} + L_{\Delta_x}  (1+\epsilon) ) e^{-\frac{c_3}{2 c_2} \Delta \tau_k }  $, $\Delta \tau_k := \tau_{k} - T_{k-1} $, and $\alpha_{st} := L_{\Delta_{st}} $.

Now, we derive the upper bound of
$| y_{st}|_{k}^- - y_{std} |$ 
with respect to the tracking error norm $\| \bx|^-_{k-1} \|$.
Because the stance foot remains static within a step, we have
	$
	y_{st}|_{k}^-  =  y_{st}|_{k-1}^+ .
	$
Then, from Eq.~\eqref{yst-bound},
	\begin{equation}
		\begin{aligned}
			| y_{st}|_{k}^+ - y_{std} | \leq 
			\| \bx|^-_{k} \| + \beta_{st}  \|  \tilde{\bx}(\tau_k;T_{k-1}^+, \bx|^+_{k-1}) \| 
			\leq  \gamma_x \| \bx|^+_{k-1} \|
		\end{aligned}
		\label{y-st}
	\end{equation}
holds, where $\gamma_x := \sqrt{\frac{c_2}{c_1}}   ( \beta_{st} +  (1+\epsilon) ) e^{-\frac{c_3}{2 c_2} \Delta \tau_k} $.
	
Finally, combining Eqs.~\eqref{c5-eq17}, \eqref{x-plus}, and \eqref{y-st} provides the following approximation of the post-impact value of the Lyapunov function $V_a$:
	\begin{equation*}
		\begin{aligned}
			V_a|^+_k &= V|^+_k + \sigma ( y_{st}|_{k}^+ - y_{std} )^2
			\leq
			c_{2}  \| \bx|^+_{k} \|^2    + \sigma   ( y_{st}|_{k}^+ - y_{std} )^2
			\\
			& \leq
			B (	c_1 \| \bx|^+_{k-1} \| ^2 +  \sigma   ( y_{st}|_{k-1}^+ - y_{std} )^2)
			\\
			&\leq
			B  (	 V|^+_{k-1} +  \sigma   ( y_{st}|_{k-1}^+ - y_{std} )^2  )
			\leq B V_a|^+_{k-1}, 
		\end{aligned}
	\end{equation*}
	where $B:=\max(  	\frac{2	c_2 \alpha_x^2 +  \sigma \gamma_x^2}{c_1},  \frac{2 c_2 \alpha_{st}}{\sigma}  )$.
	
\noindent \textbf{Evolution of $V_a$ for the hybrid model.} 
If the PD gains and $\sigma$ are chosen such that 
\begin{equation}
\tfrac{2	c_2 \alpha_x^2 +  \sigma \gamma_x^2}{c_1} < 1 
\label{c5-B}
~\mbox{and}~
\tfrac{2 c_2 \alpha_{st}}{\sigma} < 1
\end{equation}
hold (i.e., $B < 1$),
then for any $\bx|_0^+ \in B_{d_2} (\bzero)$, the sequence $\{V_a |^+_{1}$,$ V_a|^+_{2}$, $V_a|^+_{3} ... \}$ is strictly decreasing with $V_a |^+_{k} \rightarrow 0$ as $k \rightarrow \infty$.
Thus, the closed-loop hybrid system is locally asymptotically stable if the PD gains ensure that the matrix $\bA$ is Hurwitz and that Eq.~\eqref{c5-B} holds for any $\bx|_0^+ \in B_{d_2} (\bzero)$.

To meet the two inequality conditions in Eq.~\eqref{c5-B},
we can choose the function $V(\bx)$ to be $V(\bx) = \bx^T \bP \bx$ as explained in Remark~\ref{rmk-PD}.
This choice results in the continuous-phase convergence rate of $V(\bx)$ as $\frac{c_3}{c_2}=\frac{\lambda_Q}{\lambda_{max}(\bP)}$, which can be tuned with the PD gains.
Specifically, to satisfy the second inequality in Eq.~\eqref{c5-B}, we can specify $\sigma$ as any positive number such that $\sigma > 2 \lambda_{max}(\bP) \alpha_{st}$, where $\alpha_{st}$ can be estimated from system dynamics.
For instance, we can choose $\sigma$ to be $2 k_{\sigma} \lambda_{max}(\bP) \alpha_{st}$ with any constant $k_{\sigma}>1$.
Then, we can tune the PD gains to meet the first inequality in Eq.~\eqref{c5-B}, by allowing a sufficiently high continuous-phase convergence rate 
that leads to sufficiently small values of $\alpha_x$ and $\gamma_x$ for satisfying $\alpha_x^2 + \gamma_x^2 \leq \frac{c_1}{ 2 \lambda_{max}(\bP) \max (1, k_{\sigma} \alpha_{st})}$.

\noindent \textbf{Convergence of impact timings.}
When the state $\bx$ reaches zero at the steady state,
from Eq.~\eqref{c5-x_con}, the fictitious state satisfies 
$\|  \tilde{\bx}(\tau_k;T_{k-1}^+, \bx|^+_{k-1}) \|
\leq 
\sqrt{\tfrac{c_2}{c_1}} e^{-\frac{c_3}{2 c_2} (\tau_{k} - T_{k-1}) } \| \bx|^+_{k-1} \| \rightarrow 0
$
as $k \rightarrow \infty$.
Then, by Eq.~\eqref{c5-T}, $| T_k - \tau_k | \leq L_{T_x} \|  \tilde{\bx}(\tau_k;  T_{k-1}^+,\bx|^+_{k-1} )   \| \rightarrow 0$ as $k \rightarrow \infty$; that is, $T_k \rightarrow \tau_k$ as $k \rightarrow \infty$.

\noindent \textbf{Convergence of lateral foot placement.}
By the definition of $V_a$ in Eq.~\eqref{Va}, $	V_a(\bx,  y_{st} - y_{std}):=V(\bx) + \sigma  (y_{st} - y_{std})^2$, where $\sigma$ is positive and $V(\bx)$ and $(y_{st} - y_{std})^2$ are all bounded and nonnegative.
Thus, if $V_a \rightarrow 0$ as $t \rightarrow \infty$,
then $(y_{st} - y_{std})^2 \rightarrow 0$ as $t \rightarrow \infty$; that is, $y_{st} \rightarrow y_{std}$ as $t \rightarrow \infty$.

\end{document}